\documentclass[5p,times]{elsarticle}
\usepackage{amssymb}
\usepackage{amsmath}
\usepackage{lineno}
\usepackage{multirow}
\usepackage{tabularx}
\usepackage{makecell}
\usepackage{float}
\usepackage{wrapfig}
\usepackage[caption=false,font=footnotesize]{subfig}
\usepackage{graphicx}
\usepackage{adjustbox}
\usepackage{xcolor}
\usepackage{soul}
\usepackage{siunitx}
\usepackage{booktabs}
\usepackage{soul, color}
\usepackage{float}
\usepackage{multicol}
\usepackage{extdash}
\journal{Mechatronics}

\begin{document}
\begin{frontmatter}

\title{Learning-based Force Sensing and Impedance Matching for Safe Haptic Feedback in Robot-assisted Laparoscopic Surgery}

\author[aff1,aff2]{Aiden (Mohammad) Mazidi}

\author[aff3]{Majid Roshanfar\corref{cor1}}
\ead{majid.roshanfar@sickkids.ca}
\author[aff2]{Amir Sayadi}
\author[aff1]{Javad Dargahi}
\author[aff2]{Jake Barralet}
\author[aff2]{Liane S. Feldman}
\author[aff2]{Amir Hooshiar\corref{cor1}}
\ead{amir.hooshiar@mcgill.ca}

\cortext[cor1]{Corresponding authors.}

\affiliation[aff1]{organization={Surgical Robotics Laboratory (SRL), Mechanical Engineering Department, Concordia University},
  city={Montreal},
  state={QC},
  country={Canada}}

\affiliation[aff2]{organization={Surgical Performance Enhancement and Robotics (SuPER) Centre, Department of Surgery, McGill University},
  city={Montreal},
  state={QC},
  country={Canada}}
  
\affiliation[aff3]{organization={The Wilfred and Joyce Posluns Centre for Image Guided Innovation \& Therapeutic Intervention (PCIGITI), The Hospital for Sick Children (SickKids)},
  city={Toronto},
  state={ON},
  country={Canada}}

\begin{abstract}
Integrating accurate haptic feedback into robot-assisted minimally invasive surgery (RAMIS) remains challenging due to difficulties in precise force rendering and ensuring system safety during teleoperation.
We present a Nonlinear Impedance Matching Approach (NIMA) that extends our previously validated Impedance Matching Approach (IMA) by incorporating nonlinear dynamics to accurately model and render complex tool-tissue interactions in real-time.
NIMA achieves a mean absolute error of 0.01 $\pm$ 0.02 N, representing a 95\% reduction compared to IMA. Additionally, NIMA eliminates haptic ``kickback'' by ensuring zero force is applied to the user's hand when they release the handle, enhancing both patient safety and operator comfort.
By accounting for nonlinearities in tool-tissue interactions, NIMA significantly improves force fidelity, responsiveness, and precision across various surgical conditions, advancing haptic feedback systems for reliable robot-assisted surgical procedures.
\end{abstract}


\begin{highlights}
\item A novel Nonlinear Impedance Matching Approach (NIMA) for stable haptic feedback in robot-assisted surgery.
\item Neural network-based tool-tip force extraction isolating friction at the remote center of motion.
\item 95\% accuracy improvement over linear IMA with MAE of 0.01 N.
\item Elimination of haptic kickback ensuring operator safety and comfort.
\end{highlights}

\end{frontmatter}

\section{Introduction}
\label{Introduction}
Advances in haptic technology are central to enhancing patient safety, surgical precision, and operator performance in robot-assisted minimally invasive surgery (RAMIS) \cite{colan2024tactile,Amirabdollahian2017}. The integration of haptic or force feedback into teleoperated RAMIS platforms provides substantial clinical and scientific advantages, including real-time assistance to surgeons, improved perception of tissue consistency, and automatic acquisition of tissue mechanical properties~\cite{Okamura2004, torkaman2023embedded, roshanfar2025learning}. Several studies have shown that haptic feedback enhances the consistency, accuracy, and efficiency of tasks such as knot tying, thereby reducing the likelihood of inadvertent tissue damage~\cite{Mei2011}. Moreover, it improves overall surgical outcomes by enabling finer motor control and augmenting visual feedback, ultimately reducing patient risk~\cite{hooshiar2019haptic,Meijden2009,Zhou2020}. Despite these benefits, most current teleoperated surgical systems still operate without direct haptic feedback due to challenges in maintaining closed-loop stability and ensuring patient safety~\cite{Pacchierotti2016}. For instance, while next-generation RAMIS platforms such as the da Vinci~5 system have begun integrating limited force-sensing capabilities for enhanced situational awareness, they do not yet provide continuous force reflection to the operator. Consequently, the integration of robust and safe haptic feedback remains one of the most critical technological challenges in achieving full sensory immersion and automation in robotic surgery~\cite{Meijden2009,Gumbs2021}.

The regulatory approval of haptic systems as \textit{human-in-the-loop} components in RAMIS continues to face significant challenges, primarily due to the difficulty of accounting for variations in human dynamics during validation and testing~\cite{Pacchierotti2017,Okamura2004}. Stability concerns in force-feedback teleoperation and uncertainties in force estimation further complicate the safe deployment of such systems in clinical environments~\cite{Bahar2020, ghiasi2026neural}. Designing haptic devices that simultaneously ensure stability, transparency, and intuitive interaction while satisfying the diverse ergonomic and performance requirements of surgeons remains a major engineering challenge~\cite{Selim2023}. Moreover, variations in grip strength, reaction time, and motion control among individual users introduce additional uncertainties that hinder regulatory standardization and certification~\cite{Chae2018,Fu2012}. The persistent absence of high-fidelity haptic sensation in most commercial RAMIS platforms continues to limit their ability to fully replicate the tactile experience of open surgery~\cite{Lai2022}. As the adoption of robotic surgery expands globally, the integration of safe, robust, and validated haptic feedback systems (HFS) into next-generation surgical robots has become increasingly imperative to bridge this sensory gap and enhance both training and operative performance~\cite{Abiri2019}.

\subsection{Related Studies}
\label{Related Studies}
In the past decade, advances in commercial platforms and AI-enabled sensing methods have significantly advanced the provision of robust haptic feedback in RAMIS. Recent reviews and meta-analyses consistently show that augmenting vision with force or tactile information improves task performance, reduces peak forces, and can reduce errors in simulated and pre-clinical settings~\cite{patel2022haptic,bergholz2023benefits,colan2024tactile,boul2025role,dagnino2024robot}. In Table~\ref{tab:haptic_methods} we have synthesized a non-exhaustive list of current main approaches, i.e., direct tip sensing, proximal or joint-level sensing, model-based sensor-free estimation, vision-based force estimation, and purely visual or pseudo-haptic cueing, and compared them in terms of algorithmic complexity, hardware-agnostic deployment, sensor dependence, and whether they rely on proximal or distal measurements relative to the tool-tissue interface.

Instrument-integrated force sensing represents the most intuitive route to robust haptic feedback, and several recent studies have quantified its impact in realistic surgical tasks. Awad \textit{et al.} evaluated a new generation of robotic instruments with tip-mounted force sensors and demonstrated that force-feedback substantially reduced mean and maximal forces during retraction, dissection, and suturing on inanimate and ex-vivo models across experience levels~\cite{awad2024evaluation}. Systematic reviews of tactile and kinesthetic feedback in RAMIS confirm that such “true” haptics can improve precision and reduce error, but they also highlight integration complexity, sterilization constraints, and platform-specific hardware as major barriers to widespread adoption~\cite{colan2024tactile,boul2025role}. These solutions offer high-fidelity distal sensing at the cost of high mechanical and electronic complexity and low hardware agnosticity.
To avoid modifying the distal instrument, several groups have pursued proximal or sensorless approaches that reconstruct interaction forces from joint torques, actuator currents, or high-level motion descriptors. Pisla \textit{et al.} proposed an AI-based sensorless force-feedback strategy for robot-assisted esophagectomy that infers interaction forces from robot-side signals and uses them to generate haptic cues without any additional sensors on the surgical tools~\cite{pisla2025ai}. Yan \textit{et al.} developed a deep-learning model based on an ISSA-optimized backpropagation Neural Network (NN) to predict clamp forces on soft tissue using clamp motion parameters (contact area, speed, displacement, time) measured during compression tests~\cite{yan2025robust}. These methods, summarized under “sensorless/model-based” in Table~\ref{tab:haptic_methods}, tend to be more hardware-agnostic at the tool level and easier to retrofit to existing robots, but they require accurate dynamic models or extensive training data, careful calibration, and explicit handling of tissue variability to remain robust in vivo.

A complementary trend is the use of vision-based force estimation, where endoscopic images (sometimes combined with robot kinematics) are used to infer contact forces or detect over-force events. Early work by Chua \textit{et al.} combined vision and robot state to learn a mapping to interaction forces in RAMIS \cite{chua2021toward}, and subsequent studies have refined these ideas. Masui \textit{et al.} showed that deep learning models trained on laparoscopic images of a porcine kidney can estimate manipulation forces and detect over-force events when the region of interest is restricted around the tool tips~\cite{masui2024vision}. Yang \textit{et al.} introduced a contact-conditional framework that uses vision with local stiffness models and optionally joint torque information to predict forces in minimally invasive telesurgery, achieving high accuracy on contact detection and sub-10\% force prediction errors~\cite{yang2024vision}. Ding \textit{et al.} surveyed this emerging field and emphasized the potential of vision-based contact force detection to reduce hardware complexity and enable platform-agnostic haptic augmentation \cite{ding2025vision}. In Table \ref{tab:haptic_methods}, these approaches are characterized by high algorithmic complexity but minimal additional hardware, with distal sensing achieved indirectly through image analysis.

Vision-based methods naturally connect to pseudo-haptic and visual-only feedback strategies, where the goal is not to reconstruct forces precisely but to shape the surgeon’s perception of tissue stiffness or danger through visual deformation cues or other modalities. Masui \textit{et al.} explicitly frame their vision-based force estimation as an attempt to make surgeons’ existing “pseudo-haptic” impressions more explicit and actionable~\cite{masui2024vision}. Trute \textit{et al.} studied how soft-tissue visual cues in minimal-invasive and robotic surgery can support haptic perception and inform the design of visual augmentations~\cite{trute2024visual}. Broader reviews of haptic technologies for surgical simulation and robotics also stress that pseudo-haptic and multi-modal sensory substitution (e.g., vibrotactile cues) can reduce hardware demands while still improving training and performance~\cite{boul2025role}. These approaches are highly hardware-agnostic and low-cost but rely heavily on robust computer vision and careful human factor design rather than physically accurate force reconstruction.
\begin{table*}[t]
\centering
\caption{Comparison of representative studies for haptic-feedback methods in robotic surgery}
\label{tab:haptic_methods}
\footnotesize
\setlength{\tabcolsep}{3pt}
\begin{tabular}{llllll}
\toprule
\textbf{Method Class}&
\textbf{Key Works}&
\textbf{Complexity}&
\textbf{Hardware Agnosticity}&
\textbf{Sensor Dependence} &
\textbf{Prox/Distal} \\
\midrule
\textbf{Distal force sensing} &
\cite{awad2024evaluation,colan2024tactile,bergholz2023benefits} &
High (H)&
Low (L)&
High (H)&
Distal (D)\\
\textbf{Proximal/Joint-space sensing} &
\cite{pisla2025ai,yan2025robust,patel2022haptic} &
Med (M) &
Med (M) &
Robot-side only (R) &
Proximal (P)\\
\textbf{Vision-based force estimation} &
\cite{chua2021toward,masui2024vision,yang2024vision,ding2025vision} &
High (H) &
High (H) &
Low physical (L) + High data (H\textsubscript{d}) &
Distal (D\textsubscript{ind}) \\
\textbf{Pseudo-haptics / visual cues} &
\cite{masui2024vision,trute2024visual,boul2025role} &
Low--Med (L--M) &
Very High (VH) &
Low (L) &
Visual distal (V)  \\
\textbf{Hybrid multi-modal fusion} &
\cite{patel2022haptic,ding2025vision,boul2025role,dagnino2024robot} &
High (H) &
Med (M) &
Mixed (Mx) &
P + D \\
\bottomrule
\end{tabular}
\end{table*}
Taken together, recent literature suggests that there is no single “best” method for robust haptic feedback, instead, different clinical, regulatory, and engineering constraints naturally map to different regions of the design space. For new robotic platforms where instrument redesign is feasible and the cost per-unit is acceptable, instrument-integrated force sensors provide the most direct, distal measurements and the strongest evidence of performance benefits~\cite{awad2024evaluation,colan2024tactile,bergholz2023benefits}. For retrofitting legacy systems or ensuring cross-platform deployability, proximal sensing and sensorless learning-based models offer a pragmatic compromise between hardware-agnostic deployment and robustness, especially when trained on data that span the expected range of tasks and tissues~\cite{pisla2025ai,yan2025robust,patel2022haptic}. Vision-based and pseudo-haptic methods, finally, are attractive when hardware changes are impossible (e.g., closed commercial systems) or when force information is primarily needed for training, monitoring, or safety alerts rather than for continuous kinesthetic feedback~\cite{masui2024vision,yang2024vision,ding2025vision,trute2024visual}.

More recently, Impedance Matching Approach (IMA) for force feedback has emerged as an effective strategy for achieving stable and transparent haptic interaction in RAMIS. It operates as a form of sensory substitution, enabling indirect force rendering by bypassing actuators on the leader side and excluding human components from the control loop~\cite{Yin2018,Munawar2016,Prattichizzo2012}. Unlike conventional direct force feedback methods, where force error directly commands actuator output, IMA employs closed-loop position control, converting the force error into a corresponding position error that drives the robot \cite{golahmadi2021tool,haouchine2018vision,Wilfinger1994}. This formulation allows precise regulation of tool\textendash{}tissue forces while maintaining system stability through an inner position-control loop \cite{Siciliano1996}. Previous studies have validated the IMA framework through simulation and experimental evaluations, demonstrating its potential for stable and accurate haptic rendering \cite{Parsi2023}. Complementary work has explored integrating tactile object recognition and purposeful haptic exploration to extract object features and improve force perception during teleoperation \cite{Pezzementi2011}. Moreover, advanced haptic rendering algorithms have been shown to minimize feedback oscillations and enhance the fidelity of force cues, leading to more realistic and responsive interaction~\cite{Liu2022}. Building on this, the linear IMA method has recently demonstrated high precision and stability for single-axis force components in surgical teleoperation tasks~\cite{sayadi2020impedance}. We subsequently introduced a preliminary Nonlinear IMA (NIMA) framework~\cite{mazidi2024nonlinear} that accounts for nonlinear impedance parameters in tool-tissue interactions, demonstrating its feasibility for 3D force rendering with an MAE of 0.03~N and an 85\% improvement over linear IMA. However, that initial study did not address tool-tip force isolation from friction at the Remote Center of Motion (RCM), nor did it include coordinate calibration between sensing elements-limitations that the present work addresses through a neural network-based force extraction method and a comprehensive validation protocol.

\subsection{Contributions}
\label{Contributions}
\begin{figure}
\centering
\includegraphics[width=\columnwidth]{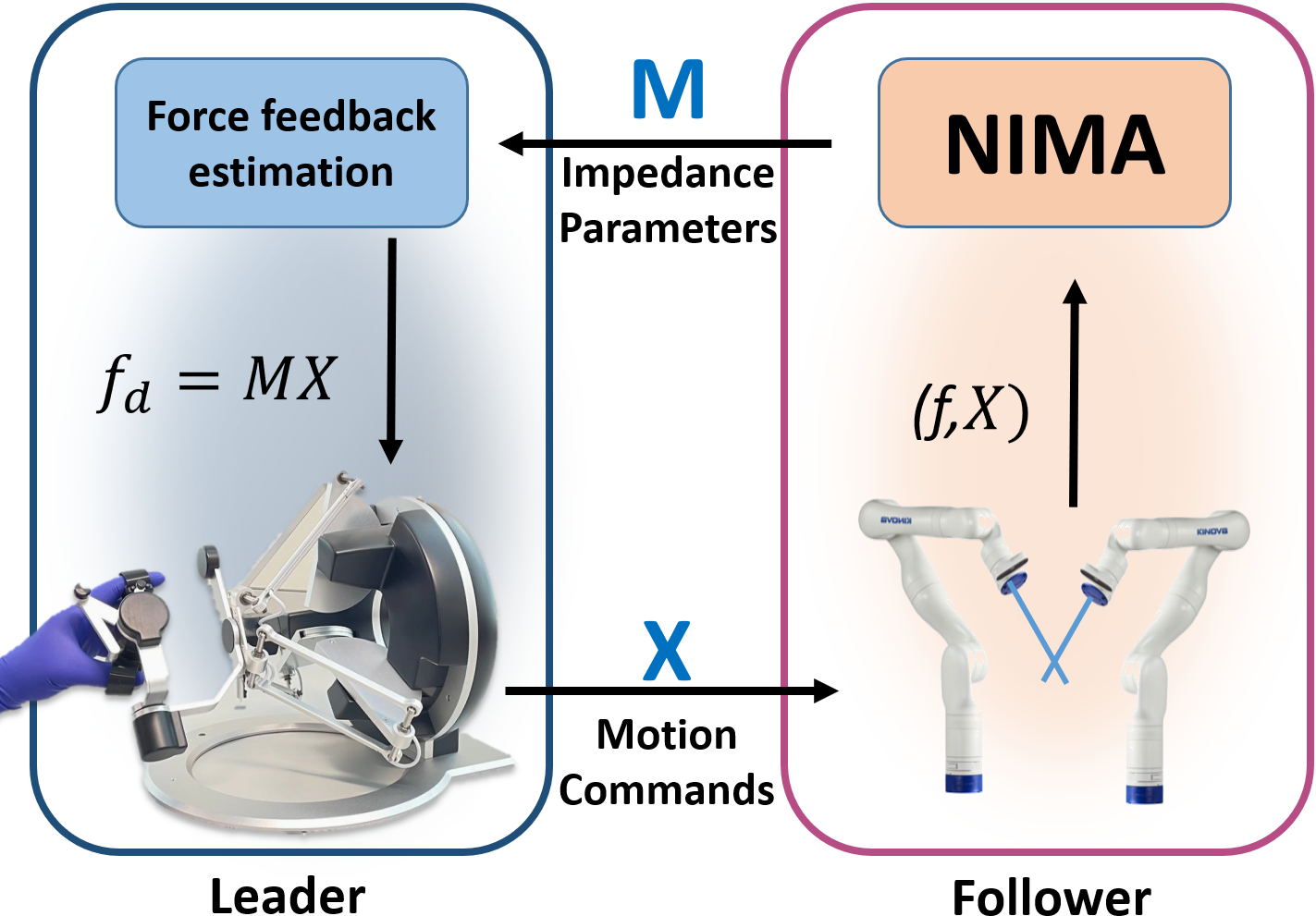}
\caption{Proposed Nonlinear Impedance Matching Approach (NIMA). The leader module (left) receives motion commands $\mathbf{X}$ from the operator and renders the estimated feedback force $\mathbf{f_d}=\mathbf{M}\mathbf{X}$, while the follower module (right) executes $\mathbf{X}$, measures $(\mathbf{f},\mathbf{X})$, and identifies nonlinear impedance parameters $\mathbf{M}$ in real time. This closed-loop structure enables stable, high-fidelity force feedback in robotic laparoscopy~\cite{mazidi2024nonlinear}.}
\label{fig:Introduction}
\end{figure}

This study extends the NIMA framework by addressing critical gaps in tool-tip force isolation and sensor calibration. As illustrated in Fig.~\ref{fig:Introduction}, the proposed framework continuously identifies nonlinear impedance characteristics in real-time to enable precise rendering of multi-axis contact forces and stable bidirectional communication between the leader and follower modules. The framework introduces a neural network (NN)-based method for extracting true tool-tip forces by isolating frictional components at the RCM, along with a coordinate correspondence calibration procedure between the robot-mounted and tissue-mounted force sensors. The main contributions of this work are:
\begin{enumerate}
    \item A NN-based tool-tip force extraction method that isolates true instrument–tissue interaction forces from frictional artifacts at the RCM.
    \item A coordinate correspondence calibration procedure enabling accurate transformation between multiple force sensor frames.
    \item An extended NIMA framework achieving sub-centinewton force rendering accuracy, representing a 95\% improvement over linear IMA.
    \item Comprehensive three-stage experimental validation including coordinate calibration, NN-based force isolation, and NIMA force reconstruction.
    \item Elimination of haptic kickback through adaptive nonlinear modeling, ensuring enhanced safety and comfort during manipulation.
\end{enumerate}

\section{Materials and Methods}
\label{Methodology}
\subsection{System Design}
\label{System Architecture}

\begin{figure*}
\centering
\begin{tabular}{cc}
\includegraphics[width=\columnwidth]{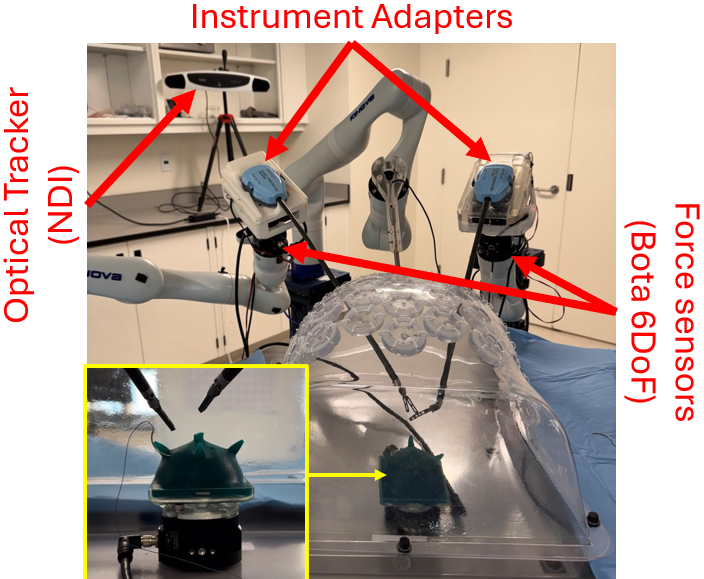} &
\includegraphics[width=0.75\columnwidth]{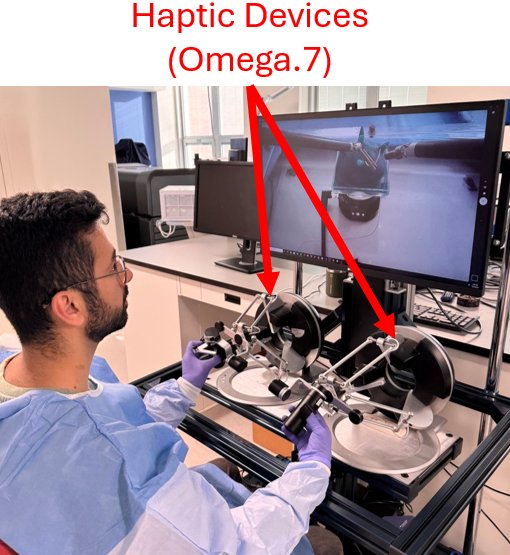} \\
(a) Follower module & (b) Leader module
\end{tabular}
\caption{
Experimental platform used to evaluate the proposed NIMA framework. 
(a) The follower module comprises two Kinova Gen3 robotic arms equipped with 6-DoF Bota force–torque sensors, custom instrument adapters, and a translucent mannequin containing a soft-tissue surrogate. (b) The leader module consists of a dual-haptic console (Omega.7, Force Dimension) used by the surgeon to teleoperate the robotic arms through motion commands, with real-time feedback displayed on a monitor.
The setup replicates a realistic minimally invasive surgical environment for evaluating tool–tissue interaction forces, motion tracking, and haptic feedback performance.}
\label{fig:Patient}
\end{figure*}

The experimental apparatus was designed to simulate a realistic surgical environment by integrating advanced technologies and components, as depicted in Fig.~\ref{fig:Patient}. The setup included a custom mannequin embedded with tissue\textendash{}representative objects, specially designed surgical tools, two Kinova Gen3 robotic arms (7~DOF), dedicated carts for robotic arm stability, two Omega.7 haptic controllers (Force Dimension), three six\textendash{}axis force\textendash{}torque sensors (SensONE, Bota Systems), and an adaptable, in-house-designed surgical console. The Kinova Gen3 robotic arms were essential for delivering precise and controlled movements of the surgical instruments, replicating the dexterity of human hands to ensure an authentic experimental environment for surgical tasks. Specialized carts were engineered to support the robotic arms, providing both stability and flexibility in positioning the system during experiments. The surgical tools were custom-designed to integrate seamlessly with the robotic arms while adhering to clinical standards and were powered by four Dynamixel actuators (ROBOTIS) to achieve fine-grained control and precision during operation. The Omega.7 haptic controllers enabled the operator to intuitively interact with the robotic arms, translating hand movements into precise surgical motions. Their high positional accuracy played a crucial role in creating realistic and user-friendly surgical simulations. A translucent mannequin, as shown in Fig.~\ref{fig:Patient}a, incorporated tissue-representative objects to replicate the structural complexities of real surgical scenarios, including pick and place and suturing tasks, creating an environment that closely mirrors actual surgical challenges. Force data from the simulated tasks were captured using Bota force sensors mounted on the tips of the robotic arms. These sensors provided valuable insights into the mechanical interactions between the surgical tools and the tissue-representative objects. The custom-designed surgical console, shown in Fig.~\ref{fig:Patient}b, acted as the central control unit, seamlessly integrating the robotic arms, haptic controllers, surgical tools, Dynamixel actuators, and force sensors, ensuring efficient communication and coordination among all components.

\begin{figure*}
\centering 
\includegraphics[width=\textwidth]{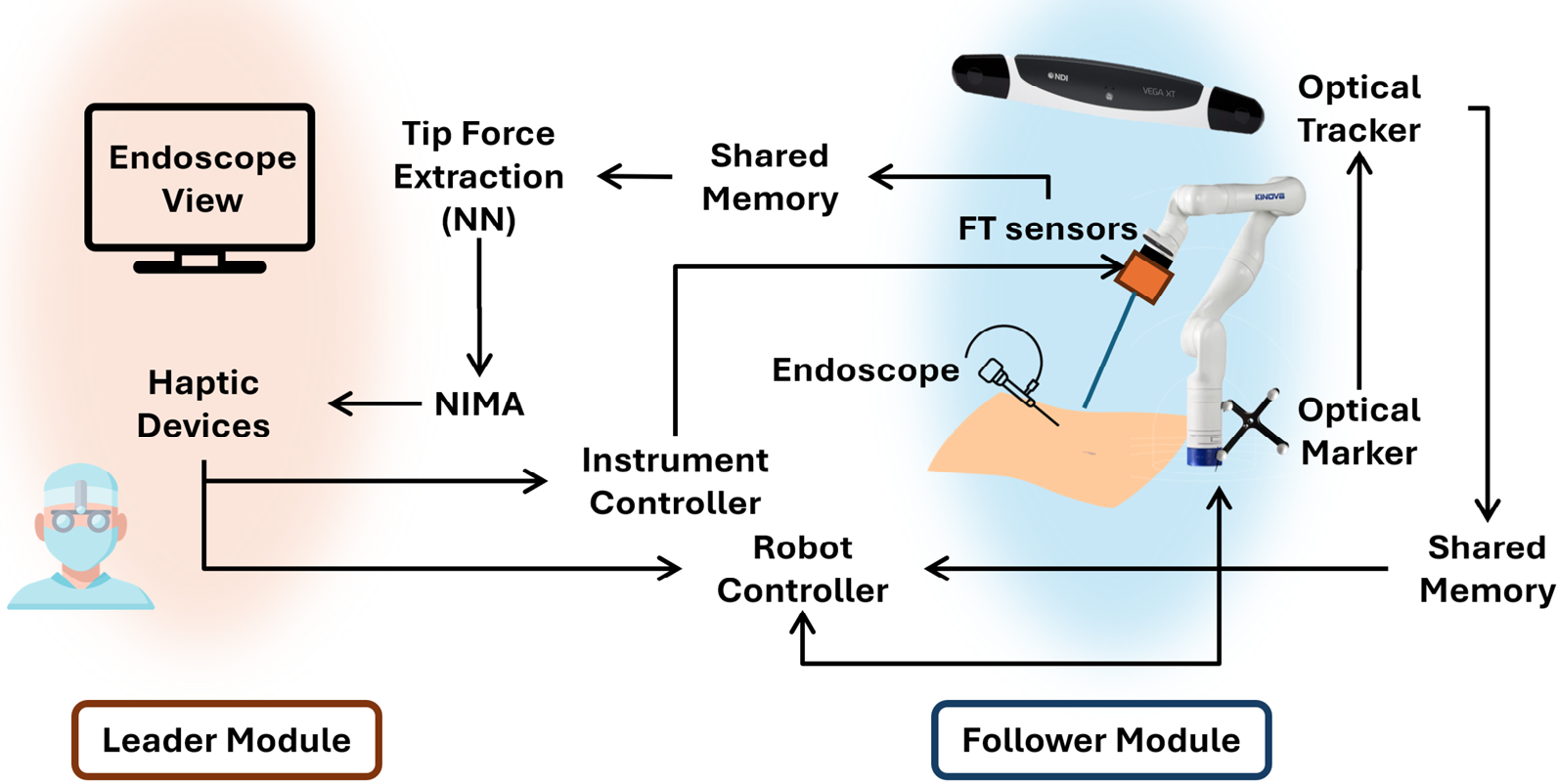} 
\caption{System architecture of the proposed NIMA-enabled robot-assisted laparoscopy setup. The leader module (left) includes dual haptic devices (Omega.7) that transmit motion commands and receive force feedback estimated by the NIMA. The follower module (right) consists of two robotic arms equipped with 6-DoF force-torque sensors, an endoscope, and optical markers tracked by an NDI system for spatial registration. Force and motion data are processed through a NN for tool-tip force extraction and used by NIMA to render stable, high-fidelity haptic.}
\label{fig:SysArch}
\end{figure*}

The system architecture, illustrated in Fig.~\ref{fig:SysArch}, enabled coordinated control by transmitting motion commands from the haptic devices to both the robot and instrument controllers. The robotic arms executed position commands, while the Dynamixel actuators controlled the orientation of the surgical tool tip, with each task managed by a dedicated control module. A PID controller was implemented for precise position and velocity regulation, ensuring accurate tool-tip alignment during operation. The robot controller consisted of two main components: one for hand–eye coordination and another for robot communication. This structure synchronized the robotic arms, optical tracker, endoscope, and haptic devices with the surgeon’s visual perspective to maintain cohesive and intuitive teleoperation. Bota 6-DoF force–torque sensors, mounted between the robots’ end effectors and the custom instrument adapters, measured the interaction forces between the surgical tools and tissue, as well as the frictional forces generated at the RCM. To maintain a stable entry point on the patient’s body, translational movements of the surgeon’s hand were converted into rotational motions about the RCM. The friction forces induced during these translational motions were calculated and subtracted from the total measured forces to isolate the true tool-tip forces. A NN module was then developed to estimate these tool-tip forces based on the signals acquired from the force sensors at the robot end effectors, as described in Section~\ref{subsec:TipForceExtraction}. The resulting tool–tissue interaction forces were transmitted to the NIMA framework, which generated adaptive haptic feedback to the surgeon’s hands, allowing the feedback intensity to be adjusted according to user preference.

For incorporating force feedback, accurately identifying the force sensor system is essential for applying gravitational biasing, which ensures precise robotic instrument movements. This study employs system identification using IMU data to determine the spatial orientation of the end effector, enabling effective gravitational biasing methods that counteract the effects of gravity on the robotic tools. To achieve this, an Integral-Free Spatial Orientation Estimation Method is introduced, which utilizes IMU sensors to calculate the end effector's orientation in real time without relying on data integration from other sources. Using this IMU data, the system effectively calculates the end effector’s roll, pitch, and yaw, critical for compensating gravitational forces acting on the robotic instruments. For simplicity, this method computes roll and pitch angles using accelerations in three directions, avoiding the integration of angular velocity over time~\cite{9397346}. Real-time angles are calculated as:
\begin{equation}
\alpha = \text{atan2}(a_y, a_z)
\end{equation}
\begin{equation}
\beta = \text{atan2}(-a_x, \sqrt{a_y^2 + a_z^2})
\end{equation}
Here, $\alpha$ represents the angle along the world's x-axis, $\beta$ is the angle along the y-axis, and $a_x$, $a_y$, and $a_z$ are acceleration values along the $x$, $y$, and $z$ directions, respectively. After deriving the spatial orientation of the end effector from the IMU data, the next step involves modeling gravitational biasing. This model is crucial for implementing techniques to counteract gravity's effects on the robotic tools. The extracted model is expressed as follows:
\begin{equation}
\textbf{AX} = \textbf{B}
\end{equation}
$\textbf{A}$ represents the system dynamics, $\textbf{X}$ is the matrix of unknown coefficients, and $\textbf{B}$ is sensor readings. The structure of $\textbf{A}$ is:
\begin{equation}
\textbf{A} =
\begin{pmatrix}
\sin\alpha_{t_1} & \cos\alpha_{t_1} & \sin\beta_{t_1} & \cos\beta_{t_1} & 1 \\
\sin\alpha_{t_2} & \cos\alpha_{t_2} & \sin\beta_{t_2} & \cos\beta_{t_2} & 1 \\
\vdots & \vdots & \vdots & \vdots & \vdots \\
\sin\alpha_{t_n} & \cos\alpha_{t_n} & \sin\beta_{t_n} & \cos\beta_{t_n} & 1
\end{pmatrix}_{n \times 5}
\end{equation}
$\textbf{X}$, the matrix of unknown coefficients, is defined as:
\begin{equation}
\textbf{X} =
\begin{pmatrix}
C_{1x} & C_{1y} & C_{1z} \\
C_{2x} & C_{2y} & C_{2z} \\
C_{3x} & C_{3y} & C_{3z} \\
C_{4x} & C_{4y} & C_{4z} \\
C_{5x} & C_{5y} & C_{5z}
\end{pmatrix}_{5 \times 3}
\end{equation}
Finally, $\textbf{B}$, containing the force readings, is expressed as:
\begin{equation}
\textbf{B} =
\begin{pmatrix}
F_{X_{t_1}} & F_{Y_{t_1}} & F_{Z_{t_1}} \\
F_{X_{t_2}} & F_{Y_{t_2}} & F_{Z_{t_2}} \\
\vdots & \vdots & \vdots \\
F_{X_{t_n}} & F_{Y_{t_n}} & F_{Z_{t_n}}
\end{pmatrix}_{n \times 3}
\end{equation}
Here, $C$ represents the unknown coefficients, and $F$ denotes the measured forces for a specific robot position. To determine the coefficient matrix $\textbf{X}$, which is fundamental to the model, the equation $\textbf{AX} = \textbf{B}$ is solved using the pseudo-inverse of $\textbf{A}$:
\begin{equation}
\textbf{X} = \textbf{A}^\dag \textbf{B}
\end{equation}
where $\textbf{A}^\dag$ is the pseudo-inverse of $\textbf{A}$, defined as:
\begin{equation}
\textbf{A}^\dag = \textbf{A}^T (\textbf{A} \textbf{A}^T)^{-1}
\end{equation}
Once the coefficient matrix is obtained, the gravitational biasing techniques can be implemented based on the extracted model from the IMU data. These techniques allow robotic instruments to counteract gravitational forces, ensuring precise and reliable movements.

\subsection{Tool-tissue Force Extraction}
\label{subsec:TipForceExtraction}
\subsubsection{Coordinate Systems Correspondence}
\begin{figure}
\centering
\includegraphics[width=\columnwidth]{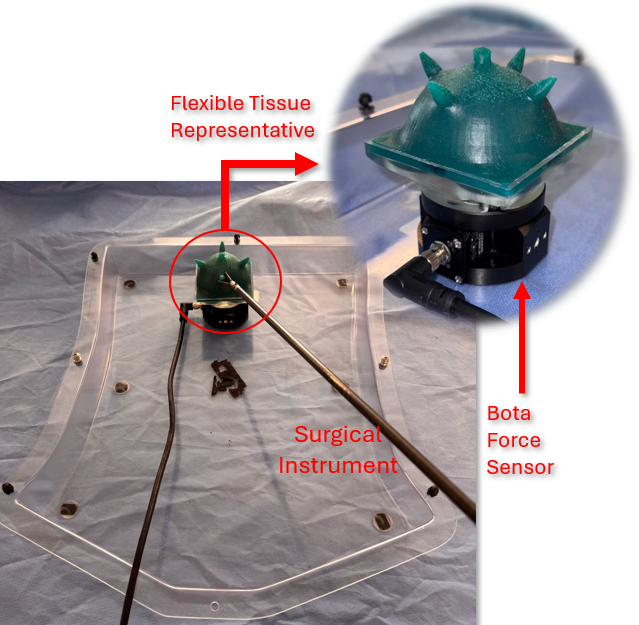}
\caption{Setup of Experiment 1, designed to determine the transformation between the coordinate systems of the two force sensors (\(S_1\) and \(S_2\)). A flexible tissue representative was mounted on a Bota 6-DoF force–torque sensor (\(S_2\)), while the second sensor (\(S_1\)) was attached to the robotic arm’s end effector. The upper section of the mannequin was removed to eliminate friction forces at the RCM, allowing the measurement of pure tool–tissue interaction forces for coordinate calibration.}
\label{fig:EXP1}
\vspace{-2 mm}
\end{figure}

After eliminating all gravitational forces due to the weight of the 3D-printed adapter, the tool-tissue forces at the tip of the surgical instrument need to be isolated from the total forces measured by the sensing element. This isolation ensures that only the tool-tip forces are rendered to the surgeon. To achieve this, two experiments were designed to gather data for training a NN capable of estimating the tool-tip forces solely using force sensors positioned outside the patient’s body. As illustrated in Fig.~\ref{fig:Patient}, the experimental setup includes two Bota FT sensors, two Kinova robotic arms (one managing the endoscope and the other controlling the surgical instrument), an instrument adaptor, a mannequin, and an NDI optical tracker. Additionally, a silicon-based flexible tissue surrogate, representing human tissue, was mounted on one of the force sensors. These experiments aimed to capture the robotic arm sensor readings, the forces at the tool tip measured via the sensor beneath the flexible tissue surrogate, and the robotic arm end-effector configuration. This data set was subsequently used to train the model.
For simplicity, we refer to the force sensor attached to the robotic arm as \( S_1 \) and the force sensor attached to the flexible tissue representative for the capture of tool-tissue forces as \( S_2 \). To use the measured tool-tip forces and those at \( S_1 \) for training, both force readings must be represented within a unified coordinate system. Thus, the first experiment aimed to transform all force readings to the coordinate system of \( S_1 \) (Fig.\ref{fig:EXP1}). In a first experiment, the upper section of the mannequin was removed to eliminate friction forces at the RCM. The robotic arm was then manipulated using haptic devices to engage the tool with the tissue surrogate. This setup ensured that \( S_1 \) only measured the interaction forces with the tissue (\( F_{r} \)), while \( S_2 \) recorded the same forces (\( F_{t} \)). Consequently, the forces \( F_{t} \), expressed in the coordinate system of \( S_{1} \), should match:
\begin{equation}
^{\{\textbf{S}_{1}\}}\textbf{f}_{r} = ^{\{\textbf{S}_{1}\}}\textbf{f}_{t}
\end{equation}
where \( ^{\{S_{1}\}}F_{r} \) represents the measured forces of \( S_1 \) expressed in its own coordinate system, and \( ^{\{S_{1}\}}F_{t} \) represents the measured forces from \( S_2 \) expressed in the coordinate system of \( S_1 \). Hence:
\begin{equation}
^{\{\textbf{S}_{1}\}}\textbf{f}_{r} = ^{\{\textbf{S}_{1}\}}\textbf{R}_{\{\textbf{S}_{2}\}} ^{\{\textbf{S}_{2}\}}\textbf{f}_{t}
\end{equation}
where the rotation matrix between the coordinate systems can be expressed as:
\begin{equation} \label{eq RotationMatrix}
^{\{\textbf{S}_{1}\}}\textbf{R}_{\{\textbf{S}_{2}\}} = ^{\{\textbf{S}_{1}\}}\textbf{R}_{\{\textbf{KBF}\}}{~}  ^{\{\textbf{KBF}\}}\textbf{R}_{\{M\}}{~}  ^{\{\textbf{M}\}}\textbf{R}_{\{\textbf{C}\}}{~}  ^{\{\textbf{C}\}}\textbf{R}_{\{\textbf{S}_{2}\}}
\end{equation}
Here, the coordinate systems are defined as follows:
\textbf{KBF} is Kinova Base Frame,
\textbf{\( S_{1} \)} is the end effector of the robot holding the instrument,
\textbf{M} is the optical marker attached to the base of the surgical instrument,
\textbf{C} is the optical tracking camera, and 
\textbf{\( S_{2} \)}is the coordinate system of the sensor measuring forces at the tip. In Eq.~\ref{eq RotationMatrix}, the only unknown matrix is \( ^{\{C\}}R_{\{S_{2}\}} \), which must be determined. This matrix can be obtained by solving an optimization problem with the objective:
\begin{equation}
    \textbf{R} \textbf{R}^T = I
\end{equation}
subject to the equality constraints:
\begin{equation}\label{Eq EulerRotation}
    \textbf{R} = \textbf{R}_Z(\theta_Z) \times \textbf{R}_Y(\theta_Y) \times \textbf{R}_X(\theta_X)
\end{equation}
\begin{equation}
    ^{\{S_{1}\}}R_{\{C\}} \times \textbf{R} \times ^{\{S_{2}\}}F_{t} -  ^{\{S_{1}\}}F_{r} = 0
\end{equation}
where \(\theta_X\), \(\theta_Y\), and \(\theta_Z\) are the Euler angles along the $X$, $Y$, and $Z$ axes, respectively. By optimizing these angles, \( ^{\{C\}}R_{\{S_{2}\}} \) can be computed using Eq.~\ref{Eq EulerRotation}.

\subsubsection{Neural Tool-Tissue Force Estimation}
\label{Sec EXP2}
After determining the required rotation matrix to map $S_1$ and $S_2$, the next step involved isolating the forces at the tool tip from those acting at the RCM. To achieve this, the upper section of the mannequin was reinstalled, and the same experimental procedure was performed on the flexible tissue representative while covering the full range of possible robot configurations and movements. By neglecting inertial effects and assuming quasi-static motion of the robotic arms, the system equations of force equilibrium on the surgical instrument were expressed as:
\begin{equation}
    \sum{\textbf{F}} = \textbf{F}_1 + \textbf{F}_2 + \textbf{F}_3 = 0
\end{equation}
where $\textbf{F}_1 \in \mathbb{R}^{1\times3}$, $\textbf{F}_2\in \mathbb{R}^{1\times3}$, and $\textbf{F}_3\in \mathbb{R}^{1\times3}$ are the forces at the robotic arm, RCM point, and tip forces. During the experiment, $\textbf{F}_1$, $\textbf{F}_3$, and quaternions of the robotic arm $\textbf{q}$ were captured. These parameters create a dataset of $30000\times 10$ for a fully connected feedforward NN designed for this regression task where $(\textbf{F}_1)_{3\times30000}$, $(\textbf{q})_{4\times30000}$ with 7 features and 30000 observations were the inputs and $(\textbf{F}_3)_{3\times30000}$ with 3 features and 30000 observations were the output.   

\begin{figure}
\centering
\includegraphics[width=\columnwidth]{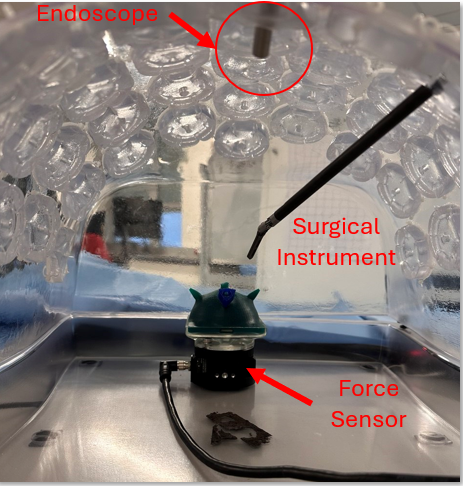}
\caption{Internal view of the mannequin used in Experiment 2, illustrating the setup for acquiring tool-tissue interaction data to train the NN model. A surgical instrument, inserted through the upper access port, interacts with a flexible tissue representative mounted on a 6-DoF Bota force–torque sensor. An endoscope provides visual feedback to monitor tool motion and contact inside the surgical workspace.}
\label{fig:EXP2}
\vspace{-2 mm}
\end{figure}

\subsection{NIMA-based Force Reconstruction}
\label{Nonlinear Impedance Matching Approach (NIMA)}
This work presents a novel NIMA as an alternative to Direct Force Reflection (DFR) for delivering force feedback in remote surgical robotics. NIMA operates by identifying nonlinear tool-tissue contact impedance parameters, denoted as $\textbf{M}$, at the follower module in real-time and transmitting these parameters to the leader module. Simultaneously, motion commands $\textbf{X}$ are sent to the follower module, where a representative laparoscopic tool interacts with a mannequin simulating soft tissue. For simplicity and generality, a polynomial nonlinear impedance model is employed to determine the NIMA parameters $\textbf{M}$. The contact force $\textbf{f}\in \mathbb{R}^{3\times1}$ is modeled as the response of a nonlinear impedance hyper-surface to the given motion command $\textbf{X}$:
\begin{eqnarray}
\label{eq:NIMA-main}
\textbf{f}=\textbf{M}\textbf{X}=\textbf{M}\begin{pmatrix}	\textbf{x}_N^{\star}&\textbf{y}_N^{\star}&\textbf{z}_N^{\star}
\end{pmatrix}^T_{1\times9N}
\end{eqnarray}
where $^T$ denotes the transpose operator, $x$, $y$, and $z$ represent the motion commands (i.e., incremental positional changes of the instrument within a time interval $\delta t$), and $_N^{\star}$ signifies the augmented state operator of degree $N$, defined as:
\begin{equation}
\textbf{u}_N^{\star}=\begin{pmatrix}
u&\dot{u}&\ddot{u}&\cdots&u^N&\dot{u}^N&\ddot{u}^N
\end{pmatrix}_{3N\times1}
\end{equation}
The NIMA impedance parameter matrix $\textbf{M}$ is structured as:
\begin{equation}
	\textbf{M}=\begin{pmatrix}
		\textbf{m}_x&\textbf{0}&\textbf{0}\\
		\textbf{0}&\textbf{m}_y&\textbf{0}\\
		\textbf{0}&\textbf{0}&\textbf{m}_z
	\end{pmatrix}_{3\times9N}
\end{equation}
where $\textbf{m}_i \in \mathbb{R}^{1\times3N}$ is a vector of impedance parameters and $\textbf{0} \in \mathbb{R}^{1\times3N}$ is a zero vector. This model assumes no cross-talk between the orthogonal tool-tissue forces along $x$, $y$, and $z$ axes. Cross-talk effects, if needed, can be modeled by including additional non-zero elements in $\textbf{M}$. To compute $\textbf{M}$ in real-time, a rolling time window of $\delta t=300$ ms was adopted. Delays under 300 ms are generally imperceptible to surgeons in leader-follower setups \cite{Perez2007}, making this an effective worst-case scenario for updating impedance parameters. The NIMA parameters $\textbf{M}$ are continuously updated using a rolling dataset of $n$ sample forces $\hat{\textbf{F}} \in \mathbb{R}^{3\times n}$ and motion commands $\hat{\textbf{X}}$:
\begin{eqnarray}
\textbf{M}=\hat{\textbf{F}}\hat{\textbf{X}}^+\\
	\label{eq:NIMA_ID}
	\hat{\textbf{F}}=\begin{pmatrix}
		\textbf{f}_{t_\circ}&\cdots&\textbf{f}_{t_\circ+\delta t}
	\end{pmatrix}_{3\times n}\\
\hat{\textbf{X}}=\begin{pmatrix}
	\textbf{X}_{t_\circ}&\cdots&\textbf{X}_{t_\circ+\delta t}
\end{pmatrix}_{9N\times n}
\end{eqnarray}
Here, $\hat{\textbf{X}}^+$ represents the pseudo-inverse of $\hat{\textbf{X}}$, computed as:
\begin{equation}
\hat{\textbf{X}}^+ = \hat{\textbf{X}}^T(\hat{\textbf{X}}\hat{\textbf{X}}^T)^{-1}
\end{equation}
The peg transfer task, a core element of the Fundamentals of Laparoscopic Surgery (FLS) curriculum~\cite{Fried2007}, was chosen for evaluation. This task involved capturing the robotic arms' positions, orientations, and velocities using two Omega.7 haptic controllers, synchronized to the operator’s motion commands at a refresh rate of 1 kHz with a programmed delay of 300 ms. Force feedback, essential for realistic haptic interaction, was recorded via force sensors (SensOne, Bota Systems) at 2 kHz. These data were used to identify NIMA parameters and apply Eq.~\ref{eq:NIMA-main} to compute and render the desired 3D force, $\textbf{f}_d$, on the haptic device. To optimize the polynomial degree $N$ for NIMA, five parallel threads with $N=1\cdots5$ were executed, with the optimal $N$ for each time window selected based on minimizing the 3D force reconstruction error. This ensured accurate force feedback and enhanced the surgical simulation's realism.

\begin{figure*}[t]
\centering
\resizebox{\textwidth}{!}{\begin{tabular}{cc}
\includegraphics[width=0.35\textwidth]{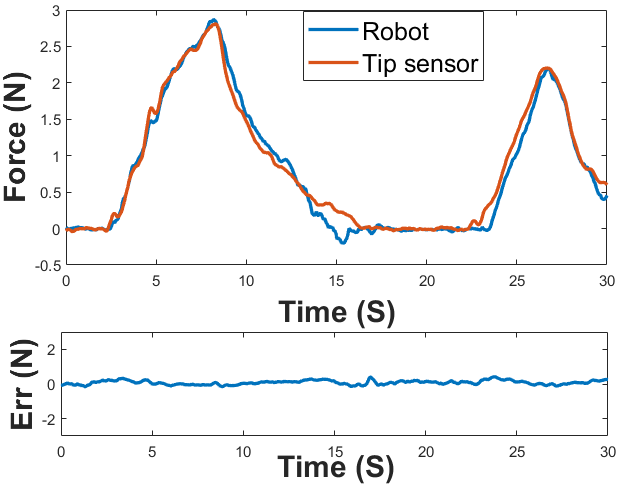} & 
\includegraphics[width=0.35\textwidth]{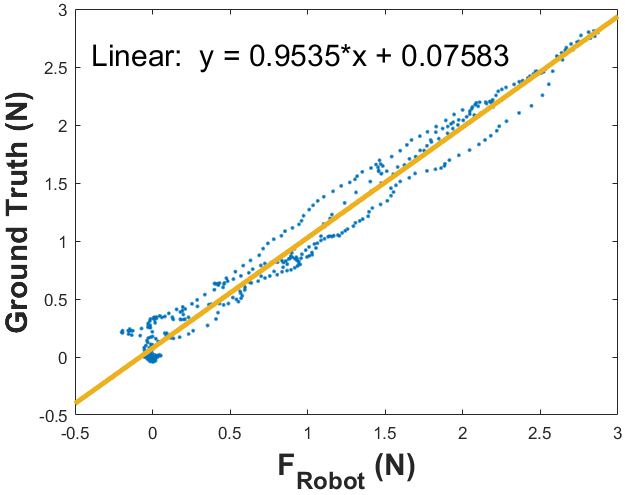} \\(a) & (b)
\end{tabular}}
\caption{(a) Comparison of forces measured by the tip sensor (ground truth) and those estimated by the robot-mounted sensor in the $Y$ direction, showing strong agreement between the two datasets. The lower subplot illustrates the corresponding force estimation error over time. (b) Linear regression analysis between the measured and estimated forces in the $Y$ direction, demonstrating a strong correlation ($R^2 = 0.95$) and validating the accuracy of the coordinate system calibration.}
\label{fig:EXP11}
\end{figure*}

\begin{figure*}[t]
\centering
\begin{tabular}{ccc}
\includegraphics[width=0.31\textwidth]{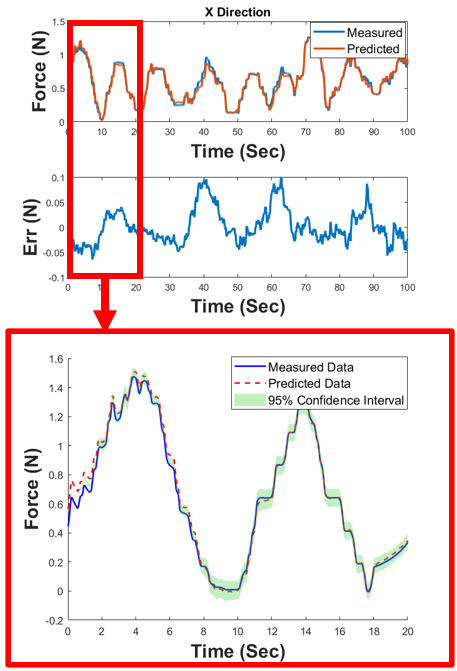} & 
\includegraphics[width=0.31\textwidth]{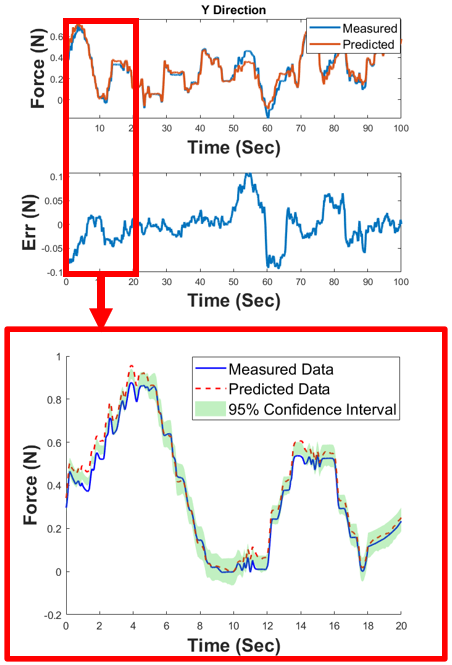} & 
\includegraphics[width=0.31\textwidth]{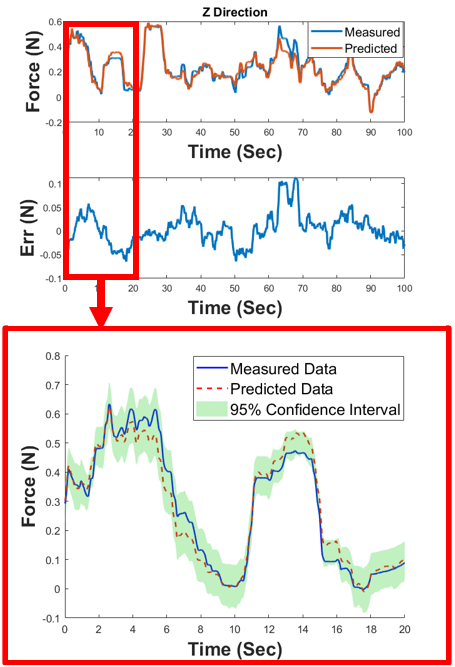} \\
(a) $X$ direction & (b) $Y$ direction & (c) $Z$ direction
\end{tabular}
\caption{Performance of the trained NN model in predicting tool-tip forces after removing frictional effects at the RCM. The plots show measured forces (ground truth) and model predictions along the (a) $X$, (b) $Y$, and (c) $Z$ directions, with shaded regions representing the 95\% confidence intervals. The close overlap between measured and predicted forces demonstrates the NN model’s high accuracy and consistency across all axes, validating its capability to capture nonlinear tool-tissue interaction dynamics.}
\label{fig:Test5}
\end{figure*}

\section{Results}
\label{Validation Studies}
This section presents a comprehensive validation of the proposed NIMA through a series of controlled experiments and performance evaluations. The results section is divided into three sections: results of Experiment 1, results of Experiment 2, and the results indicating the effectiveness of the presented NIMA model.

\begin{figure*}
\centering
\begin{tabular}{ccc}
\includegraphics[width=0.32\textwidth]{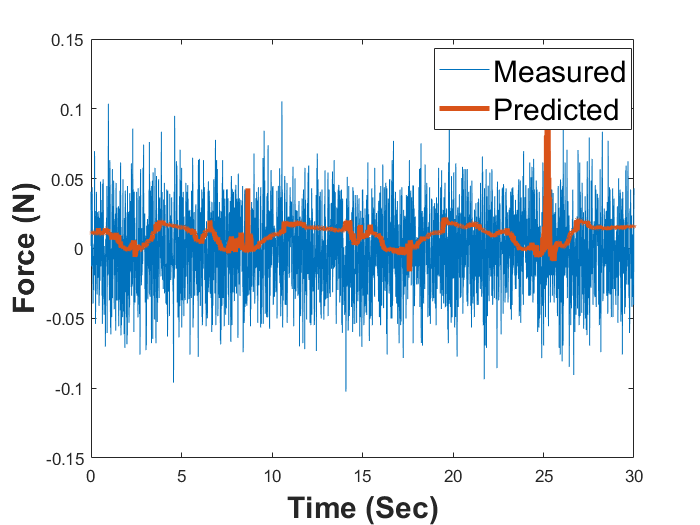} & 
\includegraphics[width=0.32\textwidth]{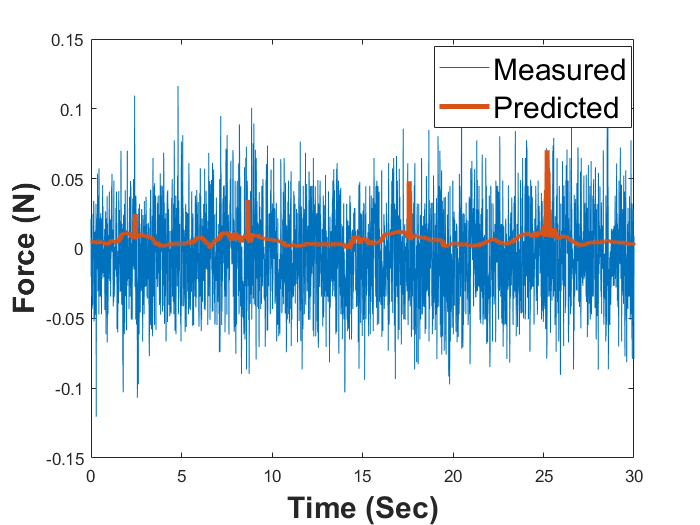} & 
\includegraphics[width=0.32\textwidth]{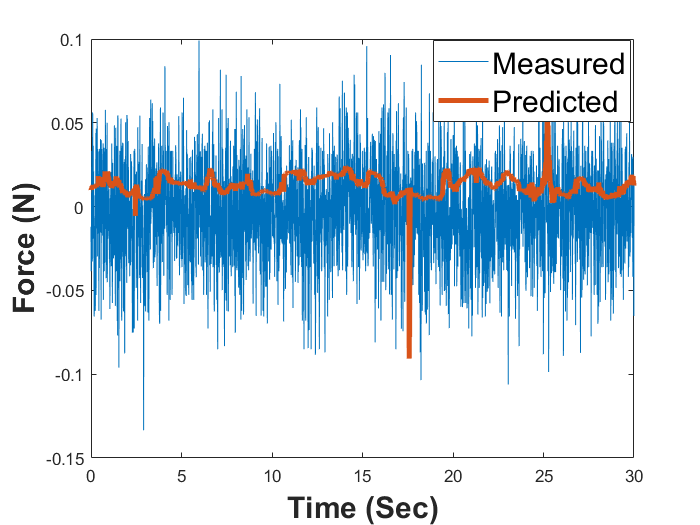} \\
(a) $X$ direction & (b) $Y$ direction & (c) $Z$ direction\\
\end{tabular}
\caption{Results of the no-contact experiment demonstrating the NN model’s capability to capture and subtract frictional forces at the RCM. Measured and predicted forces are compared along the (a) $X$, (b) $Y$, and (c) $Z$ directions. The NN model effectively suppresses sensor noise and isolates minimal residual forces, confirming its ability to eliminate frictional artifacts and ensure precise estimation of tool-tip forces.}
\label{fig:Test4}
\end{figure*}

\subsection{Validation I: Correspondence}
\label{Results of the First Experiment}
The first validation experiment, performed using the setup shown in Fig.~\ref{fig:EXP1}, evaluated whether the forces measured at the robot-mounted sensor ($S_1$) corresponded to the ground-truth tool–tissue forces from the tissue-mounted sensor ($S_2$). Removing the upper half of the mannequin eliminated RCM friction, ensuring that both sensors captured only pure interaction forces. As illustrated in Fig.~\ref{fig:EXP11}, the force traces follow the same indentation–release pattern, with both sensors exhibiting closely matched peaks, transitions, and overall waveform shape. The error signal remains tightly bounded around zero, indicating stable correspondence throughout the trial. Quantitatively, the Mean Absolute Error (MAE) between the two sensors was 0.11~N, 0.09~N, and 0.13~N for the $X$, $Y$, and $Z$ directions, with Standard Deviations (SD) values of 0.11~N, 0.12~N, and 0.18~N. The strong linear correlation in Fig.~\ref{fig:EXP11}(b) ($R^2 = 0.95$) confirms that the calibrated transformation accurately aligns the two sensor frames.

\subsection{Validation II: Tip Force Isolation}
\label{Results of the Second Experiment}
\begin{figure}
	\centering
    \includegraphics[width=\columnwidth]{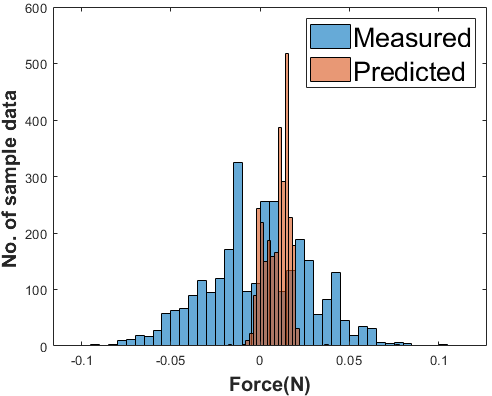}
\caption{A histogram illustrating the effectiveness of the NN model in accurately capturing forces at the RCM, facilitating the precise extraction of tool-tissue interaction forces.}
\label{fig:Test4hist}
\end{figure}

Figure~\ref{fig:Test5} presents the results of the trained NN model used to subtract friction forces at the RCM and isolate the tool-tip forces applied by the surgical instrument. In all three directions, the measured and predicted traces follow the same loading and unloading patterns, with close overlap during both rising and falling phases of the interaction. The shaded bands show the 95\% confidence intervals, within which the predicted forces remain for the vast majority of the trial. Quantitatively, the MAE between the measured and predicted forces is 0.19~N, 0.16~N, and 0.16~N for the $X$, $Y$, and $Z$ axes, respectively, with SD values of 0.2~N across all axes. These errors are mainly localized around rapid changes in force and sign reversals, while plateau regions and quasi-static segments show almost complete agreement, indicating that the NN captures the underlying tool–tissue dynamics and primarily leaves residuals associated with measurement noise and unmodeled transients. To further characterize the baseline error and verify that the NN does not introduce additional artifacts, an additional no-contact experiment was conducted (Fig.~\ref{fig:Test4}). In this configuration, the surgical instrument did not interact with the tissue surrogate, so the recorded forces represent only frictional components at the RCM and the intrinsic sensor noise. The raw sensor measurements exhibit high-frequency fluctuations, whereas the NN predictions remain close to zero throughout the trial. The mean predicted forces were 0.02~N for all three axes, with SD values of 0.03~N, showing that the model effectively suppresses noise and frictional bias. The distribution of the predicted forces along one axis, shown in Fig.~\ref{fig:Test4hist}, is tightly centered around zero, further confirming the stability of the estimator and its suitability for providing clean tool-tip force estimates for subsequent force rendering.

\begin{figure*}
	\centering
	\includegraphics[width=\textwidth]{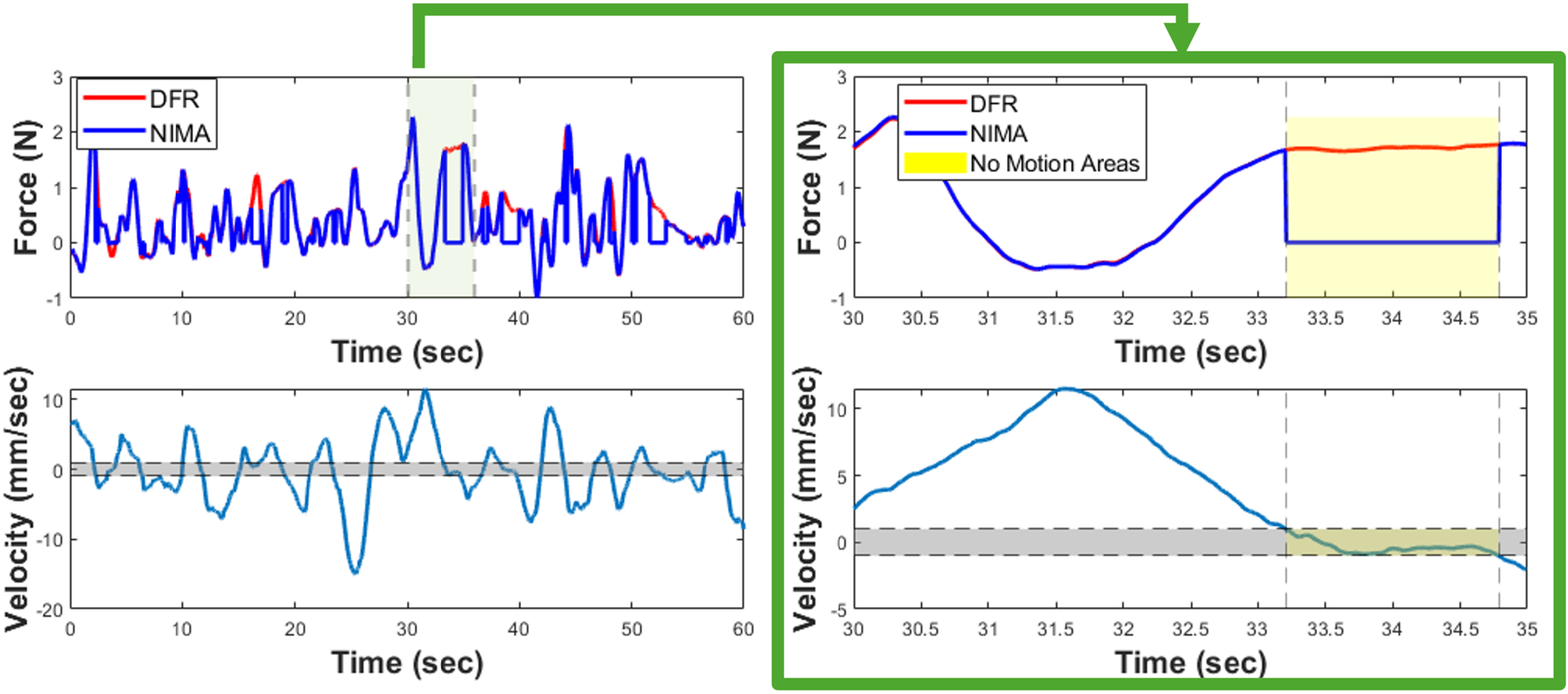}
	\caption{A comparison between the rendered force using NIMA and the measured force with DFR as the ground truth.}
	\label{fig:force_NIMA}
\end{figure*}

\subsection{Validation III: NIMA-based Force Reconstruction}
\label{Results of NIMA}
The effectiveness of our NIMA in accurately rendering forces on haptic devices is illustrated in Fig.~\ref{fig:force_NIMA}, which compares the forces rendered by the haptic device to the actual forces measured by force sensors. The implementation of NIMA achieved a MAE of 0.01~N, demonstrating high fidelity in force feedback. The errors followed a normal distribution with a SD of 0.02~N. This performance significantly exceeds that of the Linear IMA~\cite{sayadi2020impedance}, which recorded an MAE of 0.2~N and an SD of 0.4~N. A comparative assessment reveals a 95\% improvement in accuracy with NIMA. This enhancement underscores the superior precision of the nonlinear approach in generating force feedback, setting a new benchmark for realism and immersion in haptic interactions. The substantial reduction in MAE highlights the effectiveness of integrating nonlinear dynamics into impedance matching processes, improving both the quality and reliability of haptic feedback.
Furthermore, Fig.~\ref{fig:force_NIMA} demonstrates how NIMA addresses the kickback behavior by rendering no forces for movements with velocities below 1~mm/s. Consequently, when the user releases the haptic device, no force is applied to the handles, ensuring they remain stationary. However, resisting forces are rendered with high accuracy once user input exceeds the velocity threshold, preserving NIMA's precision. A detailed analysis of the algorithm's performance throughout the experiment revealed a noteworthy preference for nonlinear models, with a nonlinear fit being selected in over 64\% of the evaluated time windows. This reliance on nonlinear approaches highlights the limitations of linear impedance models in capturing the intricate, dynamic interactions between surgical tools and tissue. The complexity and variability of these interactions exceed the representational capabilities of linear models, emphasizing the nuanced nature of tool-tissue forces encountered during laparoscopic procedures. These findings suggest that temporal variations in these forces, which are crucial for realistic haptic feedback, are better captured through nonlinear modeling, enabling a more accurate and responsive simulation of surgical scenarios.

\section{Discussion}
\label{Discussion}
The experimental findings confirm that the proposed NIMA significantly enhances the stability, transparency, and realism of force feedback in RAMIS. A particularly noteworthy outcome across all trials was the complete absence of the phenomenon commonly known as the haptic kick, a sudden and undesirable jerk experienced by the operator when releasing the haptic interface. This behavior, often observed in conventional force feedback systems, typically results from the delayed response of the controller or the residual energy stored within the actuator–sensor loop. In the case of NIMA, the elimination of this artifact stems from the rapid convergence of the system’s output force vector $\textbf{X}$ to zero within a short time interval $\delta t$ once the operator releases the haptic device. Consequently, the desired force $\textbf{f}^d$ also decays smoothly to zero, ensuring that no reactive forces are transmitted back to the user. This dynamic behavior guarantees passive stability and prevents unintended force reflections, thereby improving both operator comfort and system safety. These results are consistent with the impedance-based framework previously described by Sayadi \textit{et al.}~\cite{sayadi2020impedance}, but the nonlinear formulation implemented in NIMA allows a more robust and adaptive response across a broader range of tool\textendash{}tissue interaction conditions.
Beyond mitigating haptic kickback, NIMA demonstrates strong adaptability and precision in representing nonlinear contact dynamics between the surgical instrument and the tissue phantom. Traditional linear impedance control approaches often oversimplify these interactions, limiting their ability to reproduce complex soft-tissue behaviors and variable stiffness profiles encountered during real surgical procedures. By incorporating higher\textendash{}order nonlinear terms and updating impedance parameters in real time, NIMA effectively maintains consistent performance under varying operational and material conditions. The enhanced transparency of the system allows the operator to perceive genuine tool–tissue forces without interference from control-induced artifacts, improving tactile perception during teleoperation. This capability is crucial in delicate surgical manipulations, where even small unintentional force fluctuations can lead to tissue deformation or procedural errors. The robustness and stability demonstrated by NIMA across multiple experimental conditions suggest its strong potential for integration into clinical robotic systems and surgical training simulators. By delivering stable, high-fidelity, and realistic haptic feedback, NIMA moves one step closer toward achieving the level of tactile awareness required for safe, precise, and intuitive human\textendash{}robot interaction in next-generation RAMIS platforms.

\section{Conclusions}
\label{Conclusions}
This study established NIMA as a stable and adaptable framework for delivering high-fidelity haptic feedback in RAMIS. By addressing key challenges in force rendering and system stability, NIMA provides surgeons with a realistic and safe interaction experience while maintaining smooth teleoperation. The framework integrates nonlinear impedance control with real-time force estimation, ensuring consistent and reliable haptic feedback under varying surgical conditions. Its design philosophy emphasizes simplicity, precision, and intrinsic safety, making it suitable for both clinical applications and surgical training environments.
Looking forward, several directions can further enhance NIMA’s capabilities and clinical readiness. Extending the framework to include torque feedback would provide surgeons with a richer and more intuitive sense of tool–tissue interaction, improving precision in rotational maneuvers. Integration with physics-informed neural networks (PINNs) could enable model-free, self-supervised haptic rendering, allowing the system to adapt dynamically to varying surgical environments without explicit modeling. Structural optimization of the haptic interface could ensure consistent zero-force rendering when the surgeon releases the handle, improving passivity and long-term reliability. Expanding the system to multiple degrees of freedom for combined force and torque control would make it suitable for advanced bimanual surgical tasks such as suturing or dissection.
Future work should also focus on integrating NIMA into commercial surgical robots to accelerate clinical translation and regulatory validation. Real-time optimization and clinical trials will be essential to confirm system safety and performance under operative conditions. Also, incorporating surgeon training modules and collecting user feedback will facilitate smoother adoption in both educational and clinical settings. The adaptability of NIMA also opens avenues for applications beyond laparoscopic surgery, including neurosurgical and endovascular procedures, where precise force feedback is critical for safety and efficacy. By addressing these future directions, NIMA has the potential to set a new standard for haptic feedback in robotic-assisted interventions, bridging the gap between human dexterity and robotic precision to advance the next generation of surgical robotics.

\section*{Author contributions}
A. Mazidi: conceptualization, methodology, software, validation, formal analysis, investigation, data curation, writing, original draft, writing, review and editing, visualization.  
M. Roshanfar: formal analysis, writing, review and editing.  
A. Sayadi: methodology, validation, formal analysis, writing, review and editing.  
J. Dargahi: conceptualization, resources, supervision, writing, review and editing, funding acquisition.  
J. Barralet: clinical guidance, methodology, writing, review and editing.  
L. Feldman: clinical guidance, supervision, writing, review and editing.  
A. Hooshiar: conceptualization, methodology, resources, supervision, writing, review and editing, project administration, funding acquisition.

\section*{Acknowledgments}
This research was supported by the Natural Science and Engineering Research Council (NSERC) of Canada through the NSERC CREATE Grant for Innovation at the Cutting Edge (ICE), the Fonds de Recherche du Québec pour la Nature et les Technologies (FRQNT), Concordia University, Research Institute of McGill University Health Centre (RI-MUHC), Montreal General Hospital Foundation (MGHF), and McGill University.

\section*{Ethics Statement}
This study did not involve human participants, human tissue, or animals. All experiments were conducted using phantom models (mannequins with tissue-representative surrogates). Therefore, ethics approval was not required.

\section*{Conflict of interest}
The authors declare no conflicts of interest. 

\section*{Data Availability Statement}
The data that support the findings of this study are available from the corresponding author upon reasonable request.

\section*{Permission to Reproduce Material From Other Sources}
All figures, images, and content in this paper are original works created by the authors. No material has been reproduced from other sources.

\bibliographystyle{elsarticle-num}
\bibliography{References.bib}

@article{Abiri2019,
  author = {Abiri, A. and Pensa, J. and Tao, A. and Ma, J. and Juo, Y. and Askari, S. J. and Bisley, J. W. and Rosén, J. and Dutson, E. and Grundfest, W. S.},
  title = {Multi-modal haptic feedback for grip force reduction in robotic surgery},
  journal = {Scientific Reports},
  year = {2019},
  volume = {9},
  issue = {1},
  doi = {10.1038/s41598-019-40821-1}}

@article{Amirabdollahian2017,
  author = {Amirabdollahian, F. and Livatino, S. and Vahedi, B. and Gudipati, R. and Sheen, P. and Gawrie-Mohan, S. and Vasdev, N.},
  title = {Prevalence of haptic feedback in robot-mediated surgery: a systematic review of literature},
  journal = {Journal of Robotic Surgery},
  year = {2017},
  volume = {12},
  issue = {1},
  pages = {11-25},
  doi = {10.1007/s11701-017-0763-4}}

@article{Bahar2020,
  author = {Bahar, L. and Sharon, Y. and Nisky, I.},
  title = {Surgeon-centered analysis of robot-assisted needle driving under different force feedback conditions},
  journal = {Frontiers in Neurorobotics},
  year = {2020},
  volume = {13},
  doi = {10.3389/fnbot.2019.00108}}

@article{Chae2018,
  author = {Chae, S. H. and Jung, S. and Park, H.},
  title = {In vivo biomechanical measurement and haptic simulation of portal placement procedure in shoulder arthroscopic surgery},
  journal = {Plos One},
  year = {2018},
  volume = {13},
  issue = {3},
  pages = {e0193736},
  doi = {10.1371/journal.pone.0193736}}

@article{Fried2007,
  author = {Fried, G. M.},
  title = {Fls assessment of competency using simulated laparoscopic tasks},
  journal = {Journal of Gastrointestinal Surgery},
  year = {2007},
  volume = {12},
  issue = {2},
  pages = {210-212},
  doi = {10.1007/s11605-007-0355-0}}

@article{Fu2012,
  author = {Fu, M. J. and Çavuşoğlu, M. C.},
  title = {Human-arm-and-hand-dynamic model with variability analyses for a stylus-based haptic interface},
  journal = {IEEE Transactions on Systems, Man, and Cybernetics, Part B (Cybernetics)},
  year = {2012},
  volume = {42},
  issue = {6},
  pages = {1633-1644},
  doi = {10.1109/tsmcb.2012.2197387}}

@article{golahmadi2021tool,
  title={Tool-tissue forces in surgery: A systematic review},
  author={Golahmadi, Aida Kafai and Khan, Danyal Z and Mylonas, George P and Marcus, Hani J},
  journal={Annals of Medicine and Surgery},
  volume={65},
  pages={102268},
  year={2021},
  publisher={Elsevier}}

@article{Gumbs2021,
  author = {Gumbs, A. A. and Frigerio, I. and Spolverato, G. and Croner, R. S. and Illanes, A. and Chouillard, E. and Elyan, E.},
  title = {Artificial intelligence surgery: how do we get to autonomous actions in surgery?},
  journal = {Sensors},
  year = {2021},
  volume = {21},
  issue = {16},
  pages = {5526},
  doi = {10.3390/s21165526}}

@article{haouchine2018vision,
  title={Vision-based force feedback estimation for robot-assisted surgery using instrument-constrained biomechanical three-dimensional maps},
  author={Haouchine, Nazim and Kuang, Winnie and Cotin, Stephane and Yip, Michael},
  journal={IEEE Robotics and Automation Letters},
  volume={3},
  number={3},
  pages={2160--2165},
  year={2018},
  publisher={IEEE}}

@article{hooshiar2019haptic,
	title={Haptic telerobotic cardiovascular intervention: a review of approaches, methods, and future perspectives},
	author={Hooshiar, Amir and Najarian, Siamak and Dargahi, Javad},
	journal={IEEE reviews in biomedical engineering},
	volume={13},
	pages={32--50},
	year={2019},
	publisher={IEEE}}

@article{Lai2022,
  author = {Lai, W. and Cao, L. and Liu, J. and Tjin, S. C. and Phee, S. J.},
  title = {A three-axial force sensor based on fiber bragg gratings for surgical robots},
  journal = {IEEE/ASME Transactions on Mechatronics},
  year = {2022},
  volume = {27},
  issue = {2},
  pages = {777-789},
  doi = {10.1109/tmech.2021.3071437}}

@article{Liu2022,
  author = {Liu, G. and Wang, Y. and Huang, C. and Guan, C. and Ma, D. and Wei, Z. and Qiu, X.},
  title = {Experimental evaluation on haptic feedback accuracy by using two self-made haptic devices and one additional interface in robotic teleoperation},
  journal = {Actuators},
  year = {2022},
  volume = {11},
  issue = {1},
  pages = {24},
  doi = {10.3390/act11010024}}

@article{Mei2011,
  author = {Mei, Z. and Tse, S. and Derevianko, A. and Jones, D. B. and Schwaitzberg, S. D. and Cao, C. G. L.},
  title = {Effect of haptic feedback in laparoscopic surgery skill acquisition},
  journal = {Surgical Endoscopy},
  year = {2011},
  volume = {26},
  issue = {4},
  pages = {1128-1134},
  doi = {10.1007/s00464-011-2011-8}}

@article{Meijden2009,
  author = {Meijden, O. v. d. and Schijven, M. P.},
  title = {The value of haptic feedback in conventional and robot-assisted minimal invasive surgery and virtual reality training: a current review},
  journal = {Surgical Endoscopy},
  year = {2009},
  volume = {23},
  issue = {6},
  pages = {1180-1190},
  doi = {10.1007/s00464-008-0298-x}}

@article{Munawar2016,
  author = {Munawar, A. and Fischer, G. S.},
  title = {A surgical robot teleoperation framework for providing haptic feedback incorporating virtual environment-based guidance},
  journal = {Frontiers in Robotics and AI},
  year = {2016},
  volume = {3},
  doi = {10.3389/frobt.2016.00047}}

@article{Okamura2004,
  author = {Okamura, A. M.},
  title = {Methods for haptic feedback in teleoperated robot‐assisted surgery},
  journal = {Industrial Robot: An International Journal},
  year = {2004},
  volume = {31},
  issue = {6},
  pages = {499-508},
  doi = {10.1108/01439910410566362}}

@article{Pacchierotti2016,
  author = {Pacchierotti, C. and Prattichizzo, D. and Kuchenbecker, K. J.},
  title = {Cutaneous feedback of fingertip deformation and vibration for palpation in robotic surgery},
  journal = {IEEE Transactions on Biomedical Engineering},
  year = {2016},
  volume = {63},
  issue = {2},
  pages = {278-287},
  doi = {10.1109/tbme.2015.2455932}}

@article{Pacchierotti2017,
  author = {Pacchierotti, C. and Sinclair, S. and Solazzi, M. and Frisoli, A. and Hayward, V. and Prattichizzo, D.},
  title = {Wearable haptic systems for the fingertip and the hand: taxonomy, review, and perspectives},
  journal = {IEEE Transactions on Haptics},
  year = {2017},
  volume = {10},
  issue = {4},
  pages = {580-600},
  doi = {10.1109/toh.2017.2689006}}

@article{Parsi2023,
  author = {Parsi, S. S. and Sivaselvan, M. V. and Whittaker, A. S.},
  title = {Impedance‐matching model‐in‐the‐loop simulation},
  journal = {Earthquake Engineering \&Amp; Structural Dynamics},
  year = {2023},
  volume = {52},
  issue = {12},
  pages = {3600-3621},
  doi = {10.1002/eqe.3922}}

@article{Perez2007,
  author = {Perez, M. and Quiaios, F. and Andrivon, P. and Husson, D. and Dufaut, M. and Felblinger, J. and Hubert, J.},
  title = {Paradigms and experimental set-up for the determination of the acceptable delay in telesurgery},
  journal = {2007 29th Annual International Conference of the IEEE Engineering in Medicine and Biology Society},
  year = {2007},
  doi = {10.1109/iembs.2007.4352321}}

@article{Pezzementi2011,
  author = {Pezzementi, Z. and Plaku, E. and Reyda, C. and Hager, G. D.},
  title = {Tactile-object recognition from appearance information},
  journal = {IEEE Transactions on Robotics},
  year = {2011},
  volume = {27},
  issue = {3},
  pages = {473-487},
  doi = {10.1109/tro.2011.2125350}}

@article{Prattichizzo2012,
  author = {Prattichizzo, D. and Pacchierotti, C. and Rosati, G.},
  title = {Cutaneous force feedback as a sensory subtraction technique in haptics},
  journal = {IEEE Transactions on Haptics},
  year = {2012},
  volume = {5},
  issue = {4},
  pages = {289-300},
  doi = {10.1109/toh.2012.15}}

@inproceedings{sayadi2020impedance,
  title={Impedance matching approach for robust force feedback rendering with application in robot-assisted interventions},
  author={Sayadi, Amir and Hooshiar, Amir and Dargahi, Javad},
  booktitle={2020 8th International Conference on Control, Mechatronics and Automation (ICCMA)},
  pages={18--22},
  year={2020},
  organization={IEEE}}

@article{Selim2023,
  author = {Selim, M. and Dresscher, D. and Abayazid, M.},
  title = {A comprehensive review of haptic feedback in minimally invasive robotic liver surgery: advancements and challenges},
  journal = {The International Journal of Medical Robotics and Computer Assisted Surgery},
  year = {2023},
  volume = {20},
  issue = {1},
  doi = {10.1002/rcs.2605}}

@article{Siciliano1996,
  author = {Siciliano, B.},
  title = {Parallel force/position control of robot manipulators},
  journal = {Robotics Research},
  year = {1996},
  pages = {78-89}}

@article{Wilfinger1994,
  author = {Wilfinger, L. and Wen, J. T. and Murphy, S.},
  title = {Integral force control with robustness enhancement},
  journal = {IEEE Control Systems},
  year = {1994},
  volume = {14},
  issue = {1},
  pages = {31-40},
  doi = {10.1109/37.257892}}

@article{Yin2018,
  author = {Yin, X. and Guo, S. and Song, Y.},
  title = {Magnetorheological fluids actuated haptic-based teleoperated catheter operating system},
  journal = {Micromachines},
  year = {2018},
  volume = {9},
  issue = {9},
  pages = {465},
  doi = {10.3390/mi9090465}}

@article{Zhou2020,
  author = {Zhou, M. and Qi-ming, Y. and Huang, K. and Mahov, S. and Eslami, A. and Maier, M. and Lohmann, C. P. and Navab, N. and Zapp, D. and Knoll, A. and Nasseri, M. A.},
  title = {Towards robotic-assisted subretinal injection: a hybrid parallel–serial robot system design and preliminary evaluation},
  journal = {IEEE Transactions on Industrial Electronics},
  year = {2020},
  volume = {67},
  issue = {8},
  pages = {6617-6628},
  doi = {10.1109/tie.2019.2937041}}

@ARTICLE{9397346,
  author={Hooshiar, Amir and Sayadi, Amir and Dargahi, Javad and Najarian, Siamak},
  journal={IEEE/ASME Transactions on Mechatronics}, 
  title={Integral-Free Spatial Orientation Estimation Method and Wearable Rotation Measurement Device for Robot-Assisted Catheter Intervention}, 
  year={2022},
  volume={27},
  number={2},
  pages={766-776},
  doi={10.1109/TMECH.2021.3071295}}

@article{colan2024tactile,
  title={Tactile Feedback in Robot-Assisted Minimally Invasive Surgery: A Systematic Review},
  author={Colan, Jacinto and Davila, Ana and Hasegawa, Yasuhisa},
  journal={The International Journal of Medical Robotics and Computer Assisted Surgery},
  volume={20},
  number={6},
  pages={e70019},
  year={2024},
  publisher={Wiley Online Library}
}

@article{bergholz2023benefits,
  title={The benefits of haptic feedback in robot assisted surgery and their moderators: a meta-analysis. Sci Rep 13 (1): 19215},
  author={Bergholz, M and Ferle, M and Weber, BM},
  year={2023}
}

@article{patel2022haptic,
  title={Haptic feedback and force-based teleoperation in surgical robotics},
  author={Patel, Rajni V and Atashzar, S Farokh and Tavakoli, Mahdi},
  journal={Proceedings of the IEEE},
  volume={110},
  number={7},
  pages={1012--1027},
  year={2022},
  publisher={IEEE}
}

@article{boul2025role,
  title={Role of haptic feedback technologies and novel engineering developments for surgical training and robot-assisted surgery},
  author={Boul-Atarass, Im{\'a}n Laga and Dorado, Mercedes Rubio Manzanares and Padillo-Egu{\'\i}a, Andr{\'e}s and Racero-Moreno, Jes{\'u}s and Egu{\'\i}a-Salinas, Ignacio and Pereira-Arenas, Sheila and Jim{\'e}nez-Rodr{\'\i}guez, Rosa Mar{\'\i}a and Padillo-Ruiz, Javier},
  journal={Frontiers in Robotics and AI},
  volume={12},
  pages={1567955},
  year={2025}
}

@article{dagnino2024robot,
  title={Robot-assistive minimally invasive surgery: trends and future directions},
  author={Dagnino, Giulio and Kundrat, Dennis},
  journal={International Journal of Intelligent Robotics and Applications},
  volume={8},
  number={4},
  pages={812--826},
  year={2024},
  publisher={Springer}
}

@article{awad2024evaluation,
  title={Evaluation of forces applied to tissues during robotic-assisted surgical tasks using a novel force feedback technology},
  author={Awad, Michael M and Raynor, Mathew C and Padmanabhan-Kabana, Mika and Schumacher, Lana Y and Blatnik, Jeffrey A},
  journal={Surgical endoscopy},
  volume={38},
  number={10},
  pages={6193--6202},
  year={2024},
  publisher={Springer}
}

@article{pisla2025ai,
  title={An AI-Based Sensorless Force Feedback in Robot-Assisted Minimally Invasive Surgery},
  author={Pisla, Doina and Hajjar, Nadim Al and Rus, Gabriela and Popa, Calin and Gherman, Bogdan and Ciocan, Andra and Cailean, Andrei and Radu, Corina and Chablat, Damien and Vaida, Calin and others},
  journal={Information},
  volume={16},
  number={11},
  pages={993},
  year={2025},
  publisher={MDPI}
}

@article{yan2025robust,
  title={Robust prediction of tool-tissue interaction force using ISSA-optimized BP neural networks in robotic surgery},
  author={Yan, Yong-Li and Ren, Teng and Ding, Li and Sun, Tiansheng and Huang, Shandeng},
  journal={BMC surgery},
  volume={25},
  number={1},
  pages={368},
  year={2025},
  publisher={Springer}
}

@inproceedings{chua2021toward,
  title={Toward force estimation in robot-assisted surgery using deep learning with vision and robot state},
  author={Chua, Zonghe and Jarc, Anthony M and Okamura, Allison M},
  booktitle={2021 IEEE international conference on robotics and automation (ICRA)},
  pages={12335--12341},
  year={2021},
  organization={IEEE}
}

@article{masui2024vision,
  title={Vision-based estimation of manipulation forces by deep learning of laparoscopic surgical images obtained in a porcine excised kidney experiment},
  author={Masui, Kimihiko and Kume, Naoto and Nakao, Megumi and Magaribuchi, Toshihiro and Hamada, Akihiro and Kobayashi, Takashi and Sawada, Atsuro},
  journal={Scientific Reports},
  volume={14},
  number={1},
  pages={9686},
  year={2024},
  publisher={Nature Publishing Group UK London}
}

@article{yang2024vision,
  title={Vision-based force estimation for minimally invasive telesurgery through contact detection and local stiffness models},
  author={Yang, Shuyuan and Le, My H and Golobish, Kyle R and Beaver, Juan C and Chua, Zonghe},
  journal={Journal of Medical Robotics Research},
  volume={9},
  number={03n04},
  pages={2440008},
  year={2024},
  publisher={World Scientific}
}

@article{ding2025vision,
  title={Vision-Based Contact Force Sensing in Robotic Surgery: A Technical Review},
  author={Ding, Di and Yao, Tianliang and Wang, Haoyu and Yan, Zhifan and Sun, Xusen and Luo, Rong},
  journal={IEEE Transactions on Medical Robotics and Bionics},
  year={2025},
  publisher={IEEE}
}

@article{trute2024visual,
  title={Visual cues of soft-tissue behaviour in minimal-invasive and robotic surgery},
  author={Trute, Robin Julia and Alijani, Afshin and Erden, Mustafa Suphi},
  journal={Journal of Robotic Surgery},
  volume={18},
  number={1},
  pages={401},
  year={2024},
  publisher={Springer}
}

@article{torkaman2023embedded,
  title={Embedded six-dof force--torque sensor for soft robots with learning-based calibration},
  author={Torkaman, Tannaz and Roshanfar, Majid and Dargahi, Javad and Hooshiar, Amir},
  journal={IEEE Sensors Journal},
  volume={23},
  number={4},
  pages={4204--4215},
  year={2023},
  publisher={IEEE}
}

@inproceedings{roshanfar2025learning,
  title={Learning-Based Tip Contact Force Estimation for FBG-Embedded Continuum Robots},
  author={Roshanfar, Majid and Fekri, Pedram and Nguyen, Robert H and He, Changyan and Kang, Paul H and Drake, James and Diller, Eric and Looi, Thomas},
  booktitle={2025 IEEE International Conference on Robotics and Automation (ICRA)},
  pages={844--850},
  year={2025},
  organization={IEEE}
}

@inproceedings{mazidi2024nonlinear,
  title={Nonlinear Impedance Matching Approach (NIMA) for Robust Haptic Rendering During Robotic Laparoscopy Surgery},
  author={Mazidi, Aiden and Ramos, Andr{\'e}s C and Sayadi, Amir and Dargahi, Javad and Barralet, Jake and Feldman, Liane S and Hooshiar, Amir},
  booktitle={2024 10th IEEE RAS/EMBS International Conference for Biomedical Robotics and Biomechatronics (BioRob)},
  pages={1809--1814},
  year={2024},
  organization={IEEE}
}

@article{ghiasi2026neural,
  title={Neural Collision Detection for Multi-arm Laparoscopy Surgical Robots Through Learning-from-Simulation},
  author={Ghiasi, Sarvin and Roshanfar, Majid and Barralet, Jake and Feldman, Liane S and Hooshiar, Amir},
  journal={arXiv preprint arXiv:2601.15459},
  year={2026}
}

\newpage
\onecolumn
\section*{Author Biography}

\noindent
\begin{minipage}[t]{0.12\textwidth}
    \vspace{0pt}
    \includegraphics[width=\linewidth]{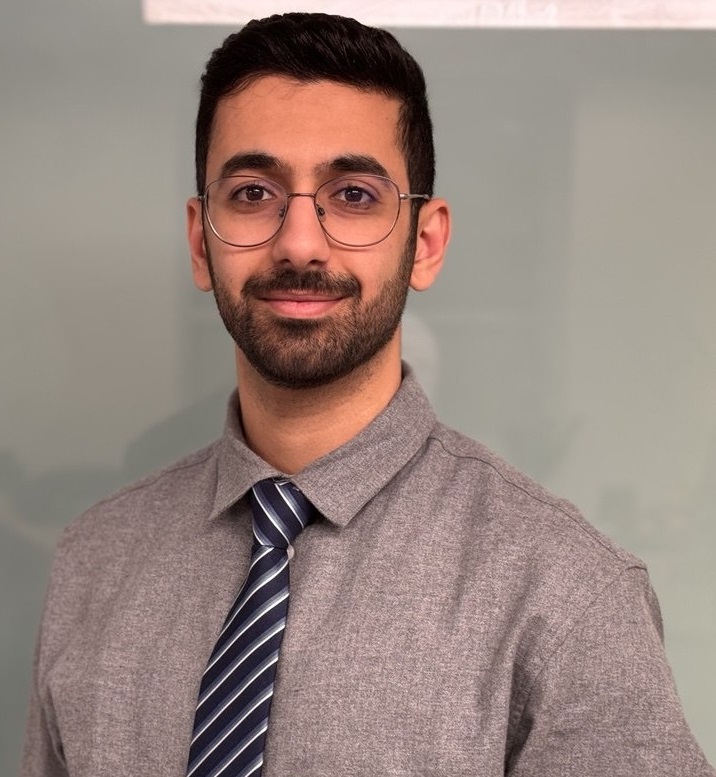}
\end{minipage}%
\hfill
\begin{minipage}[t]{0.85\textwidth}
    \vspace{0pt}
    \textbf{Aiden Mazidi} is a Master of Science candidate at Concordia University, where he serves as a Research Assistant at both McGill University's Surgical Performance Enhancement and Robotics Center (SuPER) and the Surgical Robotic Laboratory (SRL) at Concordia University. Aiden's research is focused on force rendering and haptic feedback, with a particular emphasis on RAMIS. His work involves advancing precision surgical interventions through innovative technologies in collaboration with the Surgical Innovation Program at McGill University.
\end{minipage}

\vspace{16pt}

\noindent
\begin{minipage}[t]{0.12\textwidth}
    \vspace{0pt}
    \includegraphics[width=\linewidth]{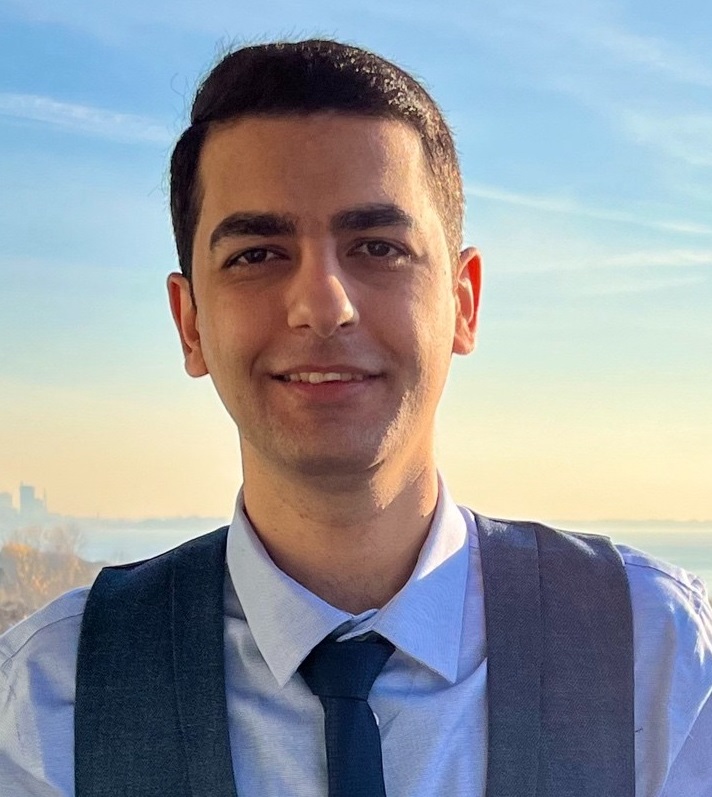}
\end{minipage}%
\hfill
\begin{minipage}[t]{0.85\textwidth}
    \vspace{0pt}
    \textbf{Majid Roshanfar} is a Postdoctoral Research Fellow at the Wilfred and Joyce Posluns Centre for Image-Guided Innovation and Therapeutic Intervention (PCIGITI) at The Hospital for Sick Children (SickKids) and the University of Toronto, Canada. He received his Ph.D. in Mechanical Engineering from Concordia University, where his research focused on hybrid-actuated soft robots with stiffness adaptation for surgical applications. His current work focuses on continuum manipulators, magnetic actuation, and force sensing for MIS.
\end{minipage}

\vspace{16pt}

\noindent
\begin{minipage}[t]{0.12\textwidth}
    \vspace{0pt}
    \includegraphics[width=\linewidth]{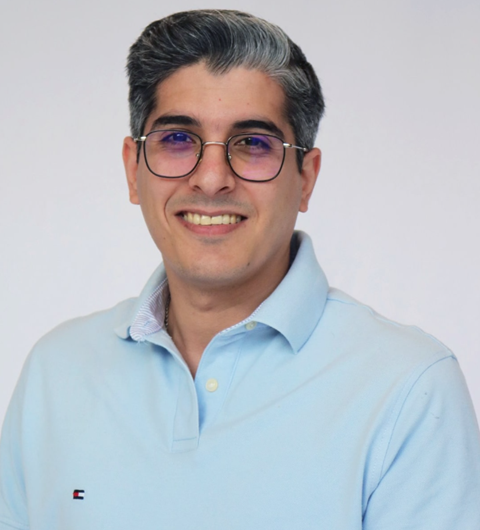}
\end{minipage}%
\hfill
\begin{minipage}[t]{0.85\textwidth}
    \vspace{0pt}
    \textbf{Amir Sayadi} is a Ph.D. candidate and researcher at the Surgical Performance Enhancement and Robotics Centre (SuPER) at McGill University, Montreal, Canada. His research focuses on medical robotics, haptic interfaces, and intelligent control systems for minimally invasive and image-guided surgical applications. His work aims to advance the development of robotic platforms that improve surgical precision.
\end{minipage}

\vspace{16pt}

\noindent
\begin{minipage}[t]{0.12\textwidth}
    \vspace{0pt}
    \includegraphics[width=\linewidth]{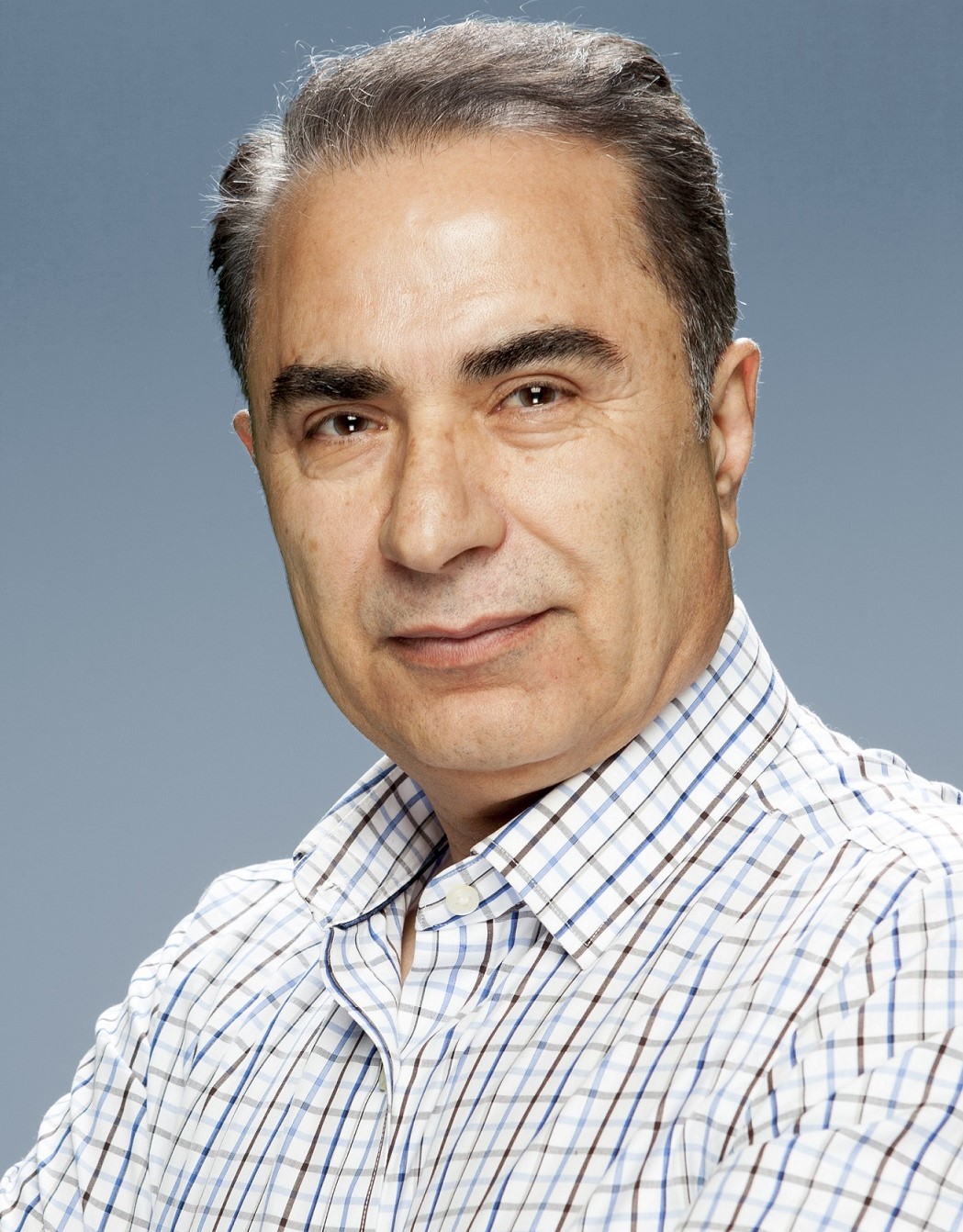}
\end{minipage}%
\hfill
\begin{minipage}[t]{0.85\textwidth}
    \vspace{0pt}
    \textbf{Javad Dargahi} received a Ph.D. degree in robotic tactile sensing from Glasgow Caledonian University. He joined the Department of Mechanical and Industrial Engineering, at Concordia University, Montreal, QC, where he is currently a Professor with the Department of Mechanical Engineering. He has co-authored three books in the scope of mechatronics and tactile sensing and has published more than 200 peer-reviewed articles. His research focuses on haptic sensors and feedback for MIS and robotics, as well as micro-manufacturing of sensors and actuators.
\end{minipage}

\vspace{16pt}

\noindent
\begin{minipage}[t]{0.12\textwidth}
    \vspace{0pt}
    \includegraphics[width=\linewidth]{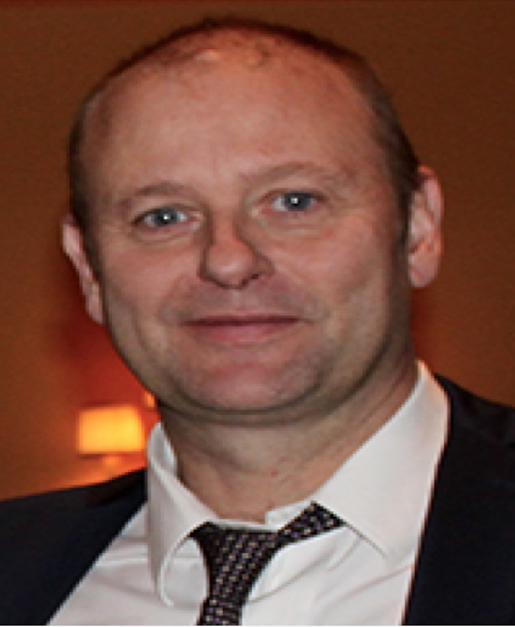}
\end{minipage}%
\hfill
\begin{minipage}[t]{0.85\textwidth}
    \vspace{0pt}
    \textbf{Jake Barralet} is currently the Vice Chair of Research and Alan Thompson Chair of Surgical Research in the Department of Surgery at McGill University, Montreal, Canada. He also serves as the Scientific Director of the McGill Clinical Innovation Platform and the Associate Director of the Surgical and Interventional Sciences Program at the Research Institute of the McGill University Health Centre. His research focuses on surgical innovation and material-mediated tissue regeneration, with an emphasis on the role of inorganic ions in bone and skin healing.
\end{minipage}

\vspace{16pt}

\noindent
\begin{minipage}[t]{0.12\textwidth}
    \vspace{0pt}
    \includegraphics[width=\linewidth]{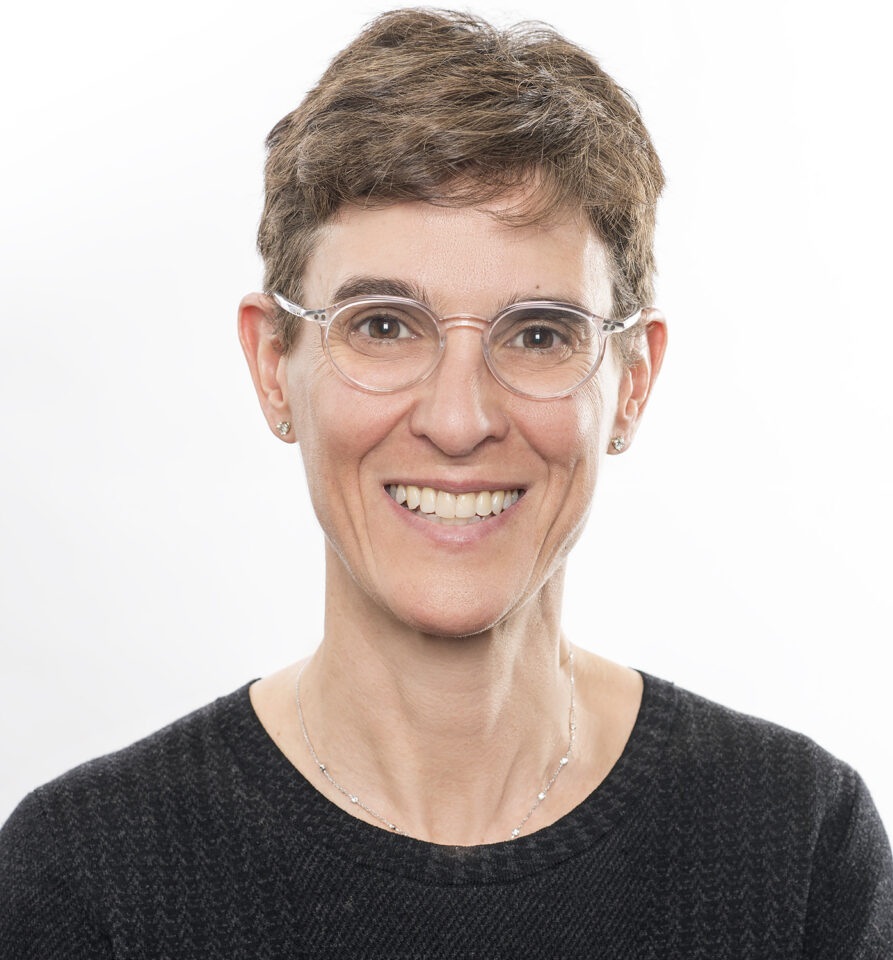}
\end{minipage}%
\hfill
\begin{minipage}[t]{0.85\textwidth}
    \vspace{0pt}
    \textbf{Liane S. Feldman} is the Edward W. Archibald Professor and Chair of Surgery at McGill University and the Surgeon-in-Chief at the McGill University Health Centre. Her clinical focus is on minimally invasive gastrointestinal surgery, and her research aims to improve recovery and outcomes following abdominal operations. She leads a multidisciplinary team implementing evidence-based Enhanced Recovery care plans recognized by Accreditation Canada. She has served as the President of the Society of American Gastrointestinal and Endoscopic Surgeons and has held leadership positions in the Canadian Association of General Surgeons and the American College of Surgeons.
\end{minipage}

\vspace{16pt}

\noindent
\begin{minipage}[t]{0.12\textwidth}
    \vspace{0pt}
    \includegraphics[width=\linewidth]{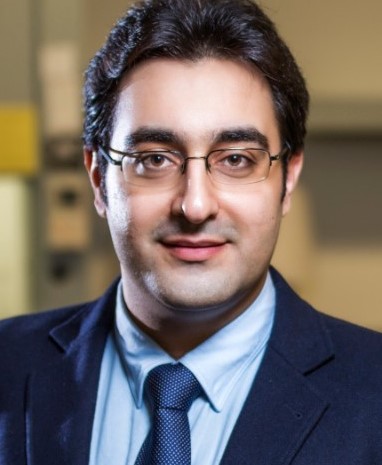}
\end{minipage}%
\hfill
\begin{minipage}[t]{0.85\textwidth}
    \vspace{0pt}
    \textbf{Amir Hooshiar} is an Assistant Professor, Edwards Distinguished Scientist, and the Director of Surgical Performance Enhancement and Robotics Centre (SuPER), at the Department of Surgery, McGill University. He specializes in the development of surgical robots and associated technologies such as surgical planning and navigation, sensing and actuation with smart materials, learning-based modelling and control of surgical robots.
\end{minipage}

\end{document}